\documentclass[journal]{IEEEtran}
\usepackage{url}
\usepackage{soul}
\usepackage{array}
\usepackage{epsfig}
\usepackage{booktabs}
\usepackage{hyperref}
\usepackage{multirow}
\usepackage{textcomp}
\usepackage{stfloats}
\usepackage{verbatim}
\usepackage{enumitem}
\usepackage{stackengine}
\usepackage{mleftright,mathtools}
\usepackage{amsthm,amsmath,amssymb,amsfonts}
\usepackage[noend]{algpseudocode}
\usepackage[comma,sort&compress,numbers]{natbib}
\usepackage[caption=false,font=normalsize,labelfont=sf,textfont=sf]{subfig}
\DeclareMathAlphabet\mathbfcal{OMS}{cmsy}{b}{n}
\newcommand{\ostar}{\mathbin{\mathpalette\make@circled\star}}
\newcommand\oast{\stackMath\mathbin{\stackinset{c}{0ex}{c}{0ex}{\ast}{\bigcirc}}}
\graphicspath{{./figures/}}

\begin{document}

\title{FCVSR: A Frequency-aware Method for Compressed Video Super-Resolution}
\author{Qiang Zhu,
Fan Zhang,~\IEEEmembership{Senior Member, IEEE,}
Feiyu Chen,~\IEEEmembership{Member, IEEE,}\\
Shuyuan Zhu,~\IEEEmembership{Member, IEEE,}
David Bull,~\IEEEmembership{Fellow, IEEE,}
and Bing Zeng,~\IEEEmembership{Fellow, IEEE}
\thanks{Q. Zhu, F. Chen, S. Zhu, and B. Zeng are with the School of Information and Communication Engineering, University of Electronic Science and Technology of China, Chengdu, China (e-mail: eezsy@uestc.edu.cn). Q. Zhu is also with the Pengcheng Laboratory, Shenzhen, China. F. Zhang and D. Bull are with the School of Computer Science, University of Bristol, Bristol, United Kingdom.}
\thanks{This work was supported by the Natural Science Foundation of Sichuan Province under Grant 2023NSFSC1972, the China Scholarship Council, the University of Bristol, and the UKRI MyWorld Strength in Places Programme (SIPF00006/1).}
}

\markboth{Journal of \LaTeX\ Class Files,~Vol.~14, No.~8, August~2021}%
{Shell \MakeLowercase{\textit{et al.}}: A Sample Article Using IEEEtran.cls for IEEE Journals}

\maketitle

\begin{abstract}
Compressed video super-resolution (SR) aims to generate high-resolution (HR) videos from the corresponding low-resolution (LR) compressed videos. Recently, some compressed video SR methods attempt to exploit the spatio-temporal information in the frequency domain, showing great promise in super-resolution performance. However, these methods do not differentiate various frequency subbands spatially or capture the temporal frequency dynamics, potentially leading to suboptimal results. In this paper, we propose a deep frequency-based compressed video SR model (FCVSR) consisting of a motion-guided adaptive alignment (MGAA) network and a multi-frequency feature refinement (MFFR) module. Additionally, a frequency-aware contrastive loss is proposed for training FCVSR, in order to reconstruct finer spatial details. The proposed model has been evaluated on three public compressed video super-resolution datasets, with results demonstrating its effectiveness when compared to existing works in terms of super-resolution performance and complexity. 
\end{abstract}

\begin{IEEEkeywords}
video super-resolution, video compression, frequency, contrastive learning, deep learning, FCVSR.
\end{IEEEkeywords}

\section{Introduction}
\IEEEPARstart{I}n recent years, video super-resolution (VSR) has become a popular research topic in image and video processing. It typically takes a low-resolution (LR) video clip and reconstructs its corresponding high-resolution (HR) counterpart with improved perceptual quality. VSR has been used for various application scenarios including video surveillance~\cite{zheng2024sac,farooq2021human}, medical imaging~\cite{chen2023cunerf,qiu2023rethinking} and video compression~\cite{kang2023super,lin2023luma}.  Inspired by the latest advances in deep learning, existing VSR methods leverage various deep neural networks~\cite{ranjan2017optical,teed2020raft,dai2017deformable,zhu2019deformable,arnab2021vivit,ho2022video} in model design, with notable examples including BasicVSR~\cite{chan2021basicvsr} and TCNet~\cite{liu2022temporal} based on optical flow~\cite{ranjan2017optical,teed2020raft}, TDAN~\cite{tian2020tdan} and EDVR~\cite{wang2019edvr} based on deformable convolution networks (DCN)~\cite{dai2017deformable, zhu2019deformable}, TTVSR~\cite{liu2022learning} and FTVSR++~\cite{qiu2023learning} based on vision transformers~\cite{arnab2021vivit}, and Upscale-A-Video~\cite{zhou2024upscale} and MGLD-VSR~\cite{yang2025motion} based on diffusion models~\cite{ho2022video}.

\begin{figure}[!t]
\centering
\centerline{\includegraphics[width=\linewidth]{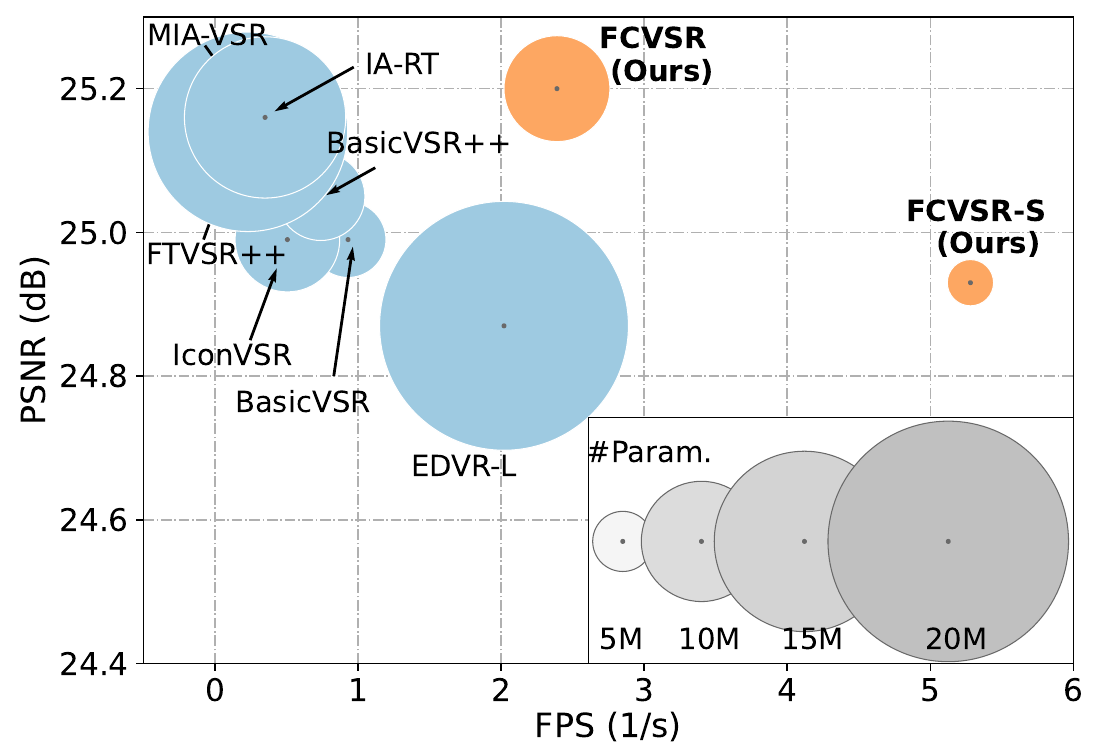}} 
\caption{Illustration of performance-complexity trade-offs for different compressed VSR models. It can be observed that the proposed FCVSR model offers better super-resolution performance with lower complexity compared to benchmark methods.}
\label{fig_perfcomplexity}
\end{figure}

When VSR is applied to video compression, it shows great potential in producing significant coding gains when integrated with conventional~\cite{afonso2018video,shen2011down} and learning-based video codecs~\cite{khani2021efficient,yang2022learned}.  In these cases, in addition to the quality degradation induced by spatial down-sampling, video compression also generates compression artifacts within low-resolution content \cite{bull2021intelligent}, which makes the super-resolution task more challenging. Previous works reported that general VSR methods may not be suitable for dealing with both compression~\cite{ding2023blind,jiang2023video,luo2022spatio} and down-sampling degradations~\cite{wang2019edvr,chan2021basicvsr}, so bespoke compressed video super-resolution methods~\cite{li2021comisr,chen2021compressed,zhang2022codec,wang2023compression,zhu2024deep,qiu2023learning,he2021fm, conde2024aim,chen2023gaussian,qiu2022learning,jiang2025compressed} have been proposed to address this issue. Among these methods, there is a class of compressed VSR models~\cite{li2021comisr,qiu2022learning,qiu2023learning} focuses on performing super-resolution in the frequency domain, such as  COMISR~\cite{li2021comisr}, FTVSR~\cite{qiu2022learning} and FTVSR++~\cite{qiu2023learning}, which align well with the nature of super-resolution, recovering the lost high-frequency details in the low-resolution content. However, it should be noted that these methods do not differentiate various frequency subbands spatially or capture the temporal frequency dynamics. This limits the reconstruction of spatial details and the accuracy of temporal alignment, resulting in suboptimal super-resolution performance.

In this context, this paper proposes a novel deep \textbf{F}requency-aware \textbf{C}ompressed \textbf{VSR} model, \textbf{FCVSR}, which exploits both spatial and temporal information in the frequency domain. It employs a new motion-guided adaptive alignment (MGAA) module that estimates multiple motion offsets between frames in the frequency domain, based on which cascaded adaptive convolutions are performed for feature alignment. We also designed a multi-frequency feature refinement (MFFR) module based on a decomposition-enhancement-aggregation strategy to restore high-frequency details within high-resolution videos. To optimize the proposed FCVSR model, we have developed a frequency-aware contrastive (FC) loss for recovering high-frequency fine details. The main contributions of this work are summarized as follows:

\begin{enumerate}[leftmargin=*]

\item A new \textbf{motion-guided adaptive alignment (MGAA)} module, which achieves improved feature alignment through explicitly considering the motion relationship in the frequency domain. 
To our knowledge, it is the first time that this type of approach is employed for video super-resolution. Compared to commonly used deformable convolution-based alignment modules~\cite{tian2020tdan,wang2019edvr,chan2022basicvsr++} in existing solutions, MGAA offers better flexibility,  higher performance, and lower complexity.

\item A novel \textbf{multi-frequency feature refinement (MFFR)} module, which provides the capability to recover fine details by using a decomposition-enhancement-aggregation strategy. Unlike existing frequency-based refinement models~\cite{xiao2024towards,li2023multi} that do not decompose features into multiple frequency subbands, our MFFR module explicitly differentiates features of different subbands, gradually performing the enhancement of subband features. 

\item  A \textbf{frequency-aware contrastive (FC)} loss is employed using contrastive learning based on the divided high-/low-frequency groups, supervising the reconstruction of finer spatial details. 

\end{enumerate}

Based on a comprehensive experiment, the proposed FCVSR model has demonstrated its superior performance in both quantitative and qualitative evaluations on three public datasets, when compared to seven existing compressed VSR methods. 
Moreover, it is also associated with relatively low computational complexity, which offers an excellent trade-off for practical applications (as shown in Fig.~\ref{fig_perfcomplexity}).

\section{Related Work} \label{RW}

This section reviews existing works in the research areas of video super-resolution (VSR), in particular focusing on compressed VSR and frequency-based VSR which are relevant to the nature of this work. We have also briefly summarized the loss functions typically used for VSR. 

\subsection{Video Super-Resolution}

VSR is a popular low-level vision task that aims to construct an HR video from its LR counterpart. State-of-the-art VSR methods~\cite{chan2021basicvsr,liu2022temporal,xiao2023online,zhu2022fffn,tian2020tdan,baniya2023omnidirectional,wang2019edvr,chan2022basicvsr++,liu2022learning,zhou2024video,xu2024enhancing,zhu2025trajectory} typically leverage various deep neural networks~\cite{ranjan2017optical,teed2020raft,dai2017deformable,zhu2019deformable,arnab2021vivit,ho2022video,tian2020tdan}, achieving significantly improved performance compared to conventional super-resolution methods based on classic signal processing theories~\cite{liu2013bayesian,xiong2010robust}. For example, 
BasicVSR~\cite{chan2021basicvsr}, IconVSR~\cite{chan2021basicvsr} and TCNet~\cite{liu2022temporal} utilize optical flow~\cite{ranjan2017optical,teed2020raft} networks to explore the temporal information between neighboring frames in order to achieve temporal feature alignment. Deformable convolution-based alignment methods~\cite{tian2020tdan, wang2019edvr} have also been proposed based on the DCN~\cite{dai2017deformable, zhu2019deformable}, with typical examples such as TDAN~\cite{tian2020tdan} and EDVR~\cite{wang2019edvr}. DCN has been reported to offer better capability in modeling geometric transformations between frames, resulting in more accurate motion estimation results. More recently, several VSR models~\cite{chan2022basicvsr++, zhu2024dvsrnet,qing2023video} have been designed with a flow-guided deformable alignment (FGDA) module that combines optical flow and DCN to achieve improved temporal alignment, among which BasicVSR++~\cite{chan2022basicvsr++} is a commonly known example. Moreover, more advanced network structures have been employed for VSR, such as Vision Transformer (ViT) and diffusion models. TTVSR~\cite{liu2022learning} is a notable ViT-based VSR method, which learns visual tokens along spatio-temporal trajectories for modeling long-range features. CTVSR~\cite{tang2023ctvsr} further exploits the strengths of Transformer-based and recurrent-based models by concurrently integrating the spatial information derived from multi-scale features and the temporal information acquired from temporal trajectories. Furthermore, diffusion models~\cite{hu2023lamd,ho2022video} have been utilized~\cite{zhou2024upscale,chen2024learning,yang2025motion} to improve the perceptual quality of super-resolved content. Examples include Upscale-A-Video~\cite{zhou2024upscale} based on a text-guided latent diffusion framework and MGLD-VSR~\cite{yang2025motion} that exploits the temporal dynamics based on a diffusion model within LR videos.

Recently, some VSR methods~\cite{qiu2022learning, qiu2023learning,li2023multi,dong2023dfvsr} are designed to perform low-resolution video up-sampling in the \textbf{frequency} domain rather than in the spatial domain. For example, FTVSR++~\cite{qiu2023learning} has been proposed to use a degradation-robust frequency-Transformer to explore the long-range information in the frequency domain; similarly, a multi-frequency representation enhancement with privilege information (MFPI) network~\cite{li2023multi} has been developed with a spatial-frequency representation enhancement branch that captures the long-range dependency in the spatial dimension, and an energy frequency representation enhancement branch to obtain the inter-channel feature relationship; DFVSR~\cite{dong2023dfvsr} applies the discrete wavelet transform to generate directional frequency features from LR frames and achieves directional frequency-enhanced alignment. Further examples include COMISR~\cite{li2021comisr} which applies a Laplacian enhancement module to generate high-frequency information for enhancing fine details, GAVSR~\cite{chen2023gaussian} that employs a high-frequency mask based on Gaussian blur to assist the attention mechanism and FTVSR~\cite{qiu2022learning} which is based on a Frequency-Transformer to conduct self-attention over a joint space-time-frequency domain. However, these frequency-based methods do not fully explore the multiple frequency subbands of the features or account for the motion relationships in the frequency domain, which restricts the exploration of more valuable information.

In many application scenarios, VSR is applied to compressed LR content, making the task even more challenging compared to uncompressed VSR. Recently, this has become a specific research focus, and numerous \textbf{compressed VSR} methods~\cite{li2021comisr,chen2021compressed,zhang2022codec,wang2023compression,zhu2024deep,qiu2023learning,he2021fm, conde2024aim,chen2023gaussian} have been developed based on coding priors. For example, CD-VSR~\cite{chen2021compressed} utilizes motion vectors, predicted frames, and prediction residuals to reduce compression artifacts and obtain spatio-temporal details for HR content; CIAF~\cite{zhang2022codec} employs recurrent models together with motion vectors to characterize the temporal relationship between adjacent frames; CAVSR~\cite{wang2023compression} also adopts motion vectors and residual frames to achieve information fusion. It is noted that these methods are typically associated with increased complexity in order to fully leverage these coding priors, which limits their adoption in practical applications.

\begin{figure*}[!t]
\centering
\includegraphics[width=0.998\textwidth]{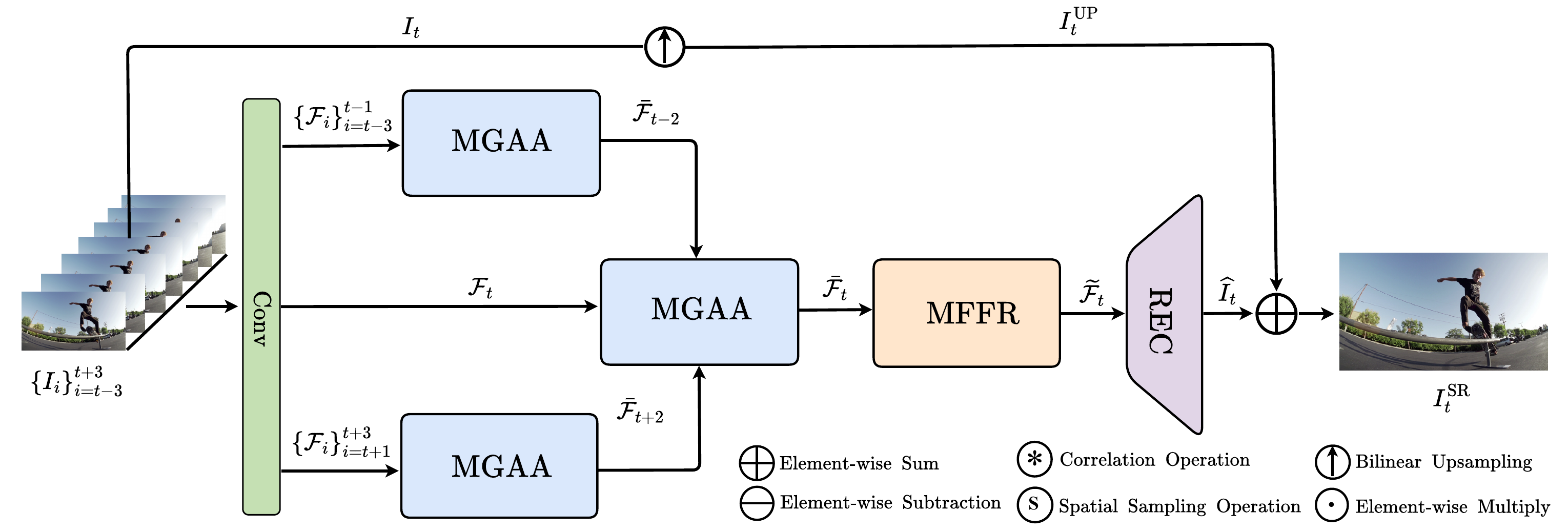}
\caption{ The architecture of the FCVSR model. A compressed LR video is fed into a convolution layer, MGAA, MFFR, and reconstruction (REC) modules to generate an HR video.}
\label{fig_pipeline}
\end{figure*}

\subsection{Loss Functions of Video Super-Resolution}

When VSR models were optimized, various loss functions were employed to address different application scenarios. 
These can be classified into two primary groups: spatial- and frequency-based. 
Spatial-based loss functions aim to minimize the pixel-wise discrepancy between the generated HR frames and the corresponding ground truth (GT) frames during training, with $L_1$ and $L_2$ losses are the commonly used objectives. Furthermore, the Charbonnier loss~\cite{lai2018fast} is a differentiable and smooth approximation of the $L_2$ loss, with similar robustness as the $L_1$ loss for reducing the weight of large errors and focusing more on smaller errors. Recently, some frequency-based loss functions~\cite{fuoli2021fourier,li2023multi,jiang2021focal} are proposed to explore the high-frequency information. For example, a Fourier space loss~\cite{fuoli2021fourier} calculated the frequency components in the Fourier domain for direct emphasis on the frequency content for restoration of high-frequency components. A focal frequency loss~\cite{jiang2021focal} generated the frequency representations using the discrete Fourier transform to supervise the generation of high-frequency information. However, these frequency-based loss functions typically observe global frequency information without decomposing features into different frequency subbands, which constrains VSR models to recover fine details.

\section{Proposed Method} \label{PM}

To address the issues associated with existing video super-resolution (VSR) methods, this paper proposes a novel frequency-aware VSR model, FCVSR, specifically for compressed content, targeting improved trade-off between performance and complexity. As illustrated in Fig.~\ref{fig_pipeline}, for the current LR video frame $I_{t}$, FCVSR takes seven LR video frames $\left\{I_{i}\right\}_{i=t-3}^{t+3}$ as input and produces an HR video frame $I_t^{\mathrm{SR}}$, targeting the uncompressed HR counterpart $I_t^{\mathrm{HR}}$ of $I_t$.

Specifically, each input frame is fed into a convolution layer with a 3$\times$3 kernel size,
\begin{equation} \label{eq_FE}
\mathcal{F}_{i}=\operatorname{Conv}\left(I_{i}\right) \in \mathbb{R}^{ h \times w \times c}, i = t-3, \dots, t+3
\end{equation}
where $h,w,c$ are the height, width, and channel of the feature. $I_{i} \in \mathbb{R}^{ h \times w \times c_I}$ and  $c_I$ is the channel of image $I_{i}$.

In order to achieve pixel-level alignment between the current frame and other input neighboring frames, multiple motion-guided adaptive alignment (MGAA) modules are employed, which take 3 sets of features generated by the convolution layer as input and outputs a single set of features. First, this is applied to the features corresponding to the first three frames, $\left\{\mathcal{F}_{i}\right\}_{i=t-3}^{t-1}$, and produces $\bar{\mathcal{F}}_{t-2}$:
\begin{equation} \label{eq_MGAA1}
\bar{\mathcal{F}}_{t-2}=\operatorname{MGAA}\left(\left\{\mathcal{F}_{i}\right\}_{i=t-3}^{t-1}\right) \in \mathbb{R}^{ h \times w \times c}.
\end{equation}
This operation is repeated for $\left\{\mathcal{F}_{i}\right\}_{i=t+1}^{t+3}$ to obtain $\bar{\mathcal{F}}_{t+2}$:
\begin{equation} \label{eq_MGAA2}
\bar{\mathcal{F}}_{t+2}=\operatorname{MGAA}\left(\left\{\mathcal{F}_{i}\right\}_{i=t+1}^{t+3}\right) \in \mathbb{R}^{ h \times w \times c}.
\end{equation}
$\bar{\mathcal{F}}_{t-2}$, $\mathcal{F}_{t}$ and $\bar{\mathcal{F}}_{t+2}$ are then fed into the MGAA module again to generate the final aligned feature set $\bar{\mathcal{F}}_{t}$:
\begin{equation} \label{eq_MGAA3}
\bar{\mathcal{F}}_{t}=\operatorname{MGAA}\left(\left\{\bar{\mathcal{F}}_{t-2}, \mathcal{F}_{t},\bar{\mathcal{F}}_{t+2}\right\}\right) \in \mathbb{R}^{ h \times w \times c}.
\end{equation}

Following the alignment operation, the aligned feature $\bar{\mathcal{F}}_{t}$ is processed by into a multi-frequency feature refinement (MFFR) module to obtain the refined feature $\widetilde{{\mathcal{F}}}_{t}$:

\begin{equation} \label{eq_MFFR}
\widetilde{\mathcal{F}}_{t}=\operatorname{MFFR}\left(\bar{\mathcal{F}}_{t}\right) \in \mathbb{R}^{ h \times w \times c}.
\end{equation}

Then, the refined feature $\widetilde{{\mathcal{F}}}_{t}$ is input into a reconstruction (REC) module to output the HR residual frame ${I}^{\mathrm{Res}}_{t}$:
\begin{equation} \label{eq_REC}
{I}^{\mathrm{Res}}_{t}=\operatorname{REC}\left(\widetilde{\mathcal{F}}_{t}\right) \in \mathbb{R}^{ h \times w \times c_I}.
\end{equation}
Finally, this is combined with the bilinear up-sampled compressed frame $I_t^{\mathrm{UP}}$ (from $I_t$) through element-wise sum to obtain the final HR frame $I_t^{\mathrm{SR}}$:

\begin{equation} \label{eq_add}
{I}^{\mathrm{SR}}_{t}={I}^{\mathrm{Res}}_{t} + I_t^{\mathrm{UP}} \in \mathbb{R}^{ h \times w \times c_I}.
\end{equation}

To facilitate the analysis, we list the notations of the major variables in Table~\ref{Tab_notions}.

\begin{table}[ht]
\caption{Notions and their description.} \label{Tab_notions}
\setlength{\tabcolsep}{2.00mm}
\fontsize{7}{9}\selectfont
\centering
\begin{tabular}{c|ccccccc}
\toprule
Notion  &  Description  \\
\midrule
$\left\{\mathcal{F}_{i}\right\}_{i=t-3}^{t-2}$ / $\left\{\mathcal{F}_{i}\right\}_{i=t-2}^{t-1}$    &   Forward / Backward feature set  
\\ 
$\bar{\mathcal{F}}_{t}$ / $\widetilde{\mathcal{F}}_{t}$     &  Aligned / Refined feature   \\ 
${I}_{t}$ / ${I}^{\mathrm{HR}}_{t}$     & Compressed LR  / Uncompressed HR frame   \\ 
${I}^{\mathrm{Res}}_{t}$ / ${I}^{\mathrm{SR}}_{t}$     &  HR residual / Reconstructed HR frame   \\ 
$\emph{\textbf{O}}_{t-2}=\left\{{o}_{n}\right\}_{n=1}^{N}$    &  Motion offsets   \\  
$\mathbf{K} = \left\{\mathbf{K}_{n}\right\}$   &   Adaptive convolution kernels   \\
$\textbf{\emph{S}}=\left\{{S}_{j}\right\}_{j=1}^Q$   &   Decomposed feature set  \\ 
$\emph{\textbf{E}} = \left\{{E}_{j} \right\}_{j=1}^Q$   &  Enhanced feature set  \\ 
$\emph{\textbf{M}}=\left\{M_{j}\right\}_{j=1}^Q$  &   Band-pass filter mask set  \\
$\mathcal{K}^{v}$ / $\mathcal{K}^{h}$   &   1-dim kernel (vertical) / (horizational) \\ 
$\bar{a}_n$  & Aligned feature  \\ 
${o}_{n}$    &   Motion offset (spatial domain)    \\ 
$\hat{o}_{n}$   &    Motion offset (frequency domain)  \\ 
$\mathcal{N}_{i}$    &   Negative set\\
$\mathcal{P}^{1}_{i}$,$\mathcal{P}^{2}_{i}$    &    Positive sets \\ 
$\mathcal{A}^{1}_{i}$, $\mathcal{A}^{2}_{i}$    &  Anchor sets  \\
\bottomrule
\end{tabular}
\end{table}

\subsection{Motion-Guided Adaptive Alignment}

\begin{figure*}[!t]
\centering
\includegraphics[width=0.998\textwidth]{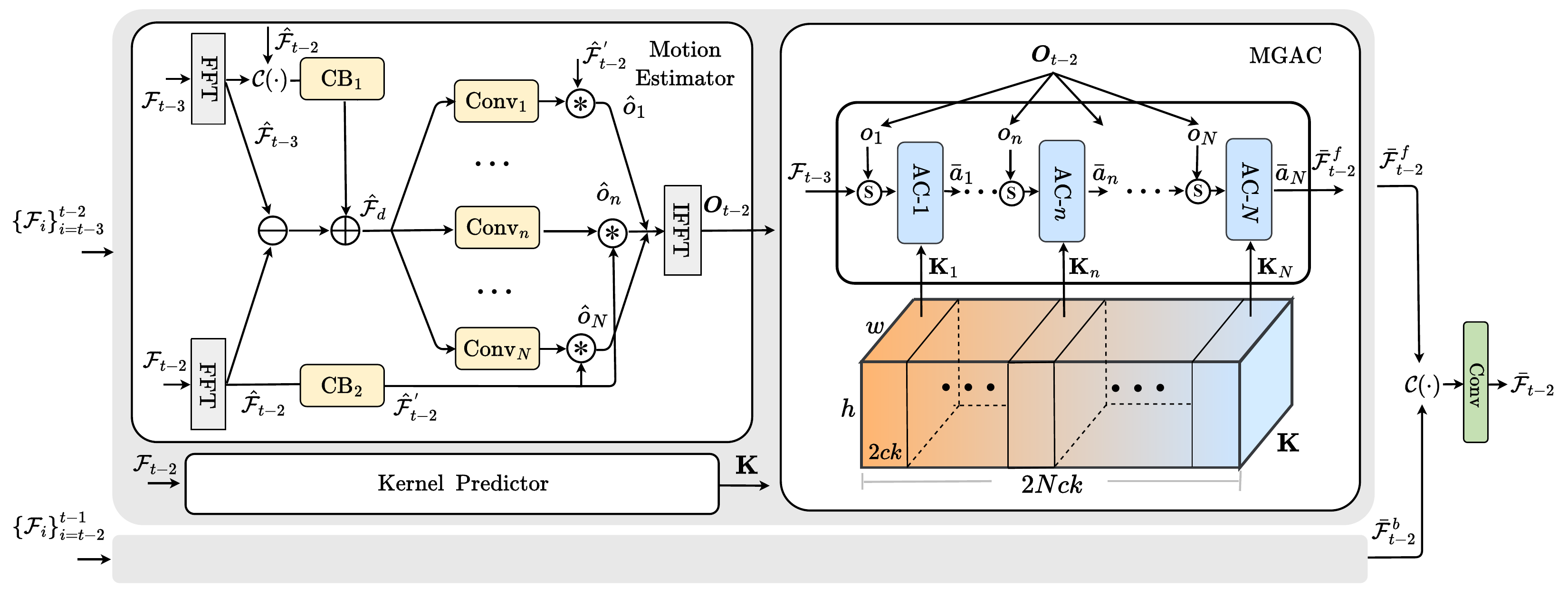} 
\caption{The architecture of motion-guided adaptive alignment (MGAA) module. The set of features $\left\{\mathcal{F}_{i}\right\}_{i=t-3}^{t-1}$ are divided into the forward set $\left\{\mathcal{F}_{i}\right\}_{i=t-3}^{t-2}$ and the backward set $\left\{\mathcal{F}_{i}\right\}_{i=t-2}^{t-1}$ for feature alignment. For the forward branch, the forward set $\left\{\mathcal{F}_{i}\right\}_{i=t-3}^{t-2}$ is first sent into the Motion Estimator (ME) module to generate the motion offsets $\emph{\textbf{O}}_{t-2} = \left\{{o}_{n}\right\}_{n=1}^{N}$. Besides, feature $\mathcal{F}_{t-2}$ is fed into the Kernel Predictor (KP) module to obtain the kernel set $\mathbf{K}=\left\{\mathbf{K}_n\right\}_{n=1}^N$. The motion offsets $\emph{\textbf{O}}_{t-2}$ and the kernel set $\mathbf{K}$ are utilized in the motion-guided adaptive convolution (MGAC) layer to achieve the feature alignment. Finally, the forwardly aligned feature $\bar{\mathcal{F}}^{f}_{t-2}$  and backwardly aligned feature $\bar{\mathcal{F}}^{b}_{t-2}$ are concatenated to obtain the output feature $\bar{\mathcal{F}}_{t-2}$.}
\label{fig_MGAA}
\end{figure*}

Most existing VSR methods estimate a single optical flow~\cite{teed2020raft, dosovitskiy2015flownet, ilg2017flownet} or offset~\cite{tian2020tdan,wang2019edvr} between frames only once to achieve feature alignment, which limits the accuracy of feature alignment in some cases. In addition, existing optical flow-based alignment modules~\cite{liu2022temporal,chan2021basicvsr,xue2019video} or deformable convolution-based alignment modules~\cite{tian2020tdan,wang2019edvr} are typically associated with high complexity, restricting their adoption in practical applications.
To address these problems, we developed a motion-guided adaptive alignment (MGAA) module that estimates different types of motion between frames, which are further used for feature alignment through adaptive convolutions. An MGAA module, as illustrated in Fig.~\ref{fig_MGAA}, consists of a Motion Estimator, Kernel Predictor, and a motion-guided adaptive convolution (MGAC) layer in a bidirectional propagation manner.

Specifically, without loss of generality, when the MGAA module takes the set of features $\left\{\mathcal{F}_{i}\right\}_{i=t-3}^{t-1}$ as input (shown in Fig.~\ref{fig_MGAA}), these features are first divided into the forward set $\left\{\mathcal{F}_{i}\right\}_{i=t-3}^{t-2}$ and the backward set $\left\{\mathcal{F}_{i}\right\}_{i=t-2}^{t-1}$ for bidirectional propagation in the MGAA module. The forward features $\left\{\mathcal{F}_{i}\right\}_{i=t-3}^{t-2}$ are then fed into the Motion Estimator $\mathrm{ME}(\cdot)$ to perform motion prediction, resulting in the set of motion offsets $\emph{\textbf{O}}_{t-2}$:
\begin{equation} \label{ME}
\emph{\textbf{O}}_{t-2} = \left\{{o}_{n}\right\}_{n=1}^{N} =  \mathrm{ME}(\mathcal{F}_{t-2},\mathcal{F}_{t-3}),  {o}_{n} \in \mathbb{R}^{ h \times w \times 2},
\end{equation}
where $N$ is the number of motion offsets.

The feature $\mathcal{F}_{t-2}$ is also input into the Kernel Predictor $\mathrm{KP}(\cdot)$ to generate $N$ adaptive convolution kernels:
\begin{equation} \label{KP}
\mathbf{K} = \left\{\mathbf{K}_{n}\right\} = \mathrm{KP}(\mathcal{F}_{t-2}), \mathbf{K}_{n} \in \mathbb{R}^{ h \times w \times 2ck},
\end{equation}
where $k$ is the kernel size of the adaptive convolution. 

Based on the motion offsets and kernel sets, the feature $\mathcal{F}_{t-3}$, is processed by the MGAC layer $\mathrm{MGAC}(\cdot)$ to achieve the feature alignment (with $\mathcal{F}_{t-2}$):
\begin{equation}
\bar{\mathcal{F}}_{t-2}^f = \mathrm{MGAC}(\mathcal{F}_{t-3}, {\emph{\textbf{O}}}_{t-2}).
\end{equation}

In parallel, the same operation is performed for the backward set to obtain the aligned features $\bar{\mathcal{F}}_{t-2}^b$. Finally, $\bar{\mathcal{F}}_{t-2}^f$ and $\bar{\mathcal{F}}_{t-2}^b$ are concatenated and fed into a convolution layer to obtain the final aligned feature $\bar{\mathcal{F}}_{t-2}$.

\subsubsection{Motion Estimator}

The existing alignment modules ~\cite{chan2021basicvsr,liu2022temporal,tian2020tdan,wang2019edvr,chan2022basicvsr++,zhu2024dvsrnet,qing2023video} only use a single optical flow or offset to achieve the feature alignment, which limits the alignment to explore the diverse motion and handle the complex motions. Recently, some works~\cite{li2023multi, zhu2023exploring} have been designed to explore motion information between frames in the frequency domain for video restoration. Such as MFPI~\cite{li2023multi} proposed a frequency-based representation alignment to capture the motion between frames for generating fine-grained details of aligned features. ETFS~\cite{zhu2023exploring} investigated that the frequency spectrum of the blurry video has more obvious temporal differences than the spectrum of the sharp video and proposed a frequency spectrum prior-guided alignment module that leverages the blur information in the frequency domain to relieve the negative impact of the unpredictable blur in the alignment. However, these methods neglect the receptive fields of motion information, which limits the diverse motion offsets generation and further accurate feature alignment.

In this work, we introduce the motion learning between frames in the frequency domain into the compressed video super-resolution and propose a Motion Estimator to explore more accurate motion information with different receptive fields between compressed frames for feature alignment. The input feature sets are first performed by the Fast Fourier Transform (FFT) to generate the frequency feature sets. Then, the real part and imaginary part of these frequency features are concatenated together along the channel dimension. Therefore, the resulting frequency features are denoted as $\mathcal{\hat{F}}_{t-2},\mathcal{\hat{F}}_{t-3}\in \mathbb{R}^{ h \times w \times 2c}$  corresponding to features $\mathcal{F}_{t-2},\mathcal{F}_{t-3}$:  
\begin{equation}
\begin{aligned}
&\mathcal{\hat{F}}_{t-2} =\mathcal{C}(\mathrm{FFT}.\texttt{Real}(\mathcal{F}_{t-2}), \mathrm{FFT}.\texttt{Imag}(\mathcal{F}_{t-2})) \\
&\mathcal{\hat{F}}_{t-3} =\mathcal{C}(\mathrm{FFT}.\texttt{Real}(\mathcal{F}_{t-3}), \mathrm{FFT}.\texttt{Imag}(\mathcal{F}_{t-3})), \\
\end{aligned}
\end{equation}
where $\mathrm{FFT}.\texttt{Real}(\cdot)$ and $\mathrm{FFT}.\texttt{Imag}(\cdot)$ are the extract operations of the real part and imaginary part of the FFT operation. $\mathcal{C}(\cdot)$ represents the concatenation operation.
To fully explore the difference between frequency features for motion estimation, we use the subtraction operation and concatenation operation (through a convolution block $\mathrm{CB}_1$) for frequency features to obtain the difference feature $\mathcal{\hat{F}}_{d}\in \mathbb{R}^{ h \times w \times 2c}$: 
\begin{equation}
\mathcal{\hat{F}}_{d} =\mathcal{\hat{F}}_{t-2} - \mathcal{\hat{F}}_{t-3} + \mathrm{CB}_{1}(\mathcal{C}(\mathcal{\hat{F}}_{t-2},\mathcal{\hat{F}}_{t-3})),
\end{equation}
where $\mathrm{CB}_{1}$ is a convolution block consisting of a 3$\times$3 convolution layer with $4c$ channels, a ReLU activation function followed  a 3$\times$3 convolution layer with $2c$ channels. 

The difference feature  $\mathcal{\hat{F}}_{d}$ is then input into multiple branches with different kernel sizes to learn motion sets $\left\{\hat{o}_{n}\right\}$ in the frequency domain. For the $n$-th branch, the motion offset is calculated as follows: 
\begin{equation}
\hat{o}_{n} = \mathrm{Conv}_\mathrm{n}(\mathcal{\hat{F}}_{d}) \oast \mathrm{CB}_2( \mathcal{\hat{F}}_{t-2})\in \mathbb{R}^{ h \times w \times 4},
\end{equation}
where $\mathrm{Conv}_\mathrm{n}$ consists of two convolution layers with kernel size $2n + 1$, a PReLU activation function, and channel attention (CA)~\cite{zhang2018image}. In $\mathrm{Conv}_\mathrm{n}$, the first convolution layer has $2c$ channels and the second convolution layer has $4$ channels. 
$\mathrm{CA}$ consists of an average pooling layer, two convolution layers with $c$ channels, and a sigmoid activation function.
$\oast$ is a correlation operation \cite{teed2020raft} to obtain the correlation between features. This correlation operation is formed by taking the dot product between all pairs of features, and the correlation can be efficiently computed as a single matrix multiplication. $\mathrm{CB}_2$ is a convolution block consisting of a 3$\times$3 convolution layer with $2c$ channels, a ReLU activation function and a 3$\times$3 convolution layer with 4 channels.

The learned multiple frequency motion offsets are transformed into the spatial domain by inverse FFT, resulting in the motion offsets $\emph{\textbf{O}}_{t-2} = \left\{{o}_{n}\right\}_{n=1}^{N}$.

\subsubsection{Kernel Predictor}

Adaptive convolution achieves better performance and lower complexity compared to the conventional convolution in some video-related tasks~\cite{liu2024motion,zhou2019spatio}. These works have validated the advantages of adaptive convolutions.
To predict adaptive convolution kernels, we designed a Kernel Predictor $\mathrm{KP}(\cdot)$ (formulated by Eq. (\ref{KP})), which consists of a 3$\times$3 convolution layer and a 1$\times$1 convolution layer to generate two directional kernels. The kernel set $\mathbf{K}$ predicted here is a $2Nck$-dim vector representing $N$ sets of kernels $\left\{\mathbf{K}_n\right\}_{n=1}^N$. For the $n$-th predicted kernel $\mathbf{K}_n$, it has two 1-dim kernels $\mathcal{K}^{v}_n$ and $\mathcal{K}^{h}_n$ with sizes $k \times 1$ and $1 \times k$ and $c$ channels, along vertical and horizational directions, respectively.

\begin{figure*}[!t]
\centering
\includegraphics[width=0.97\textwidth]{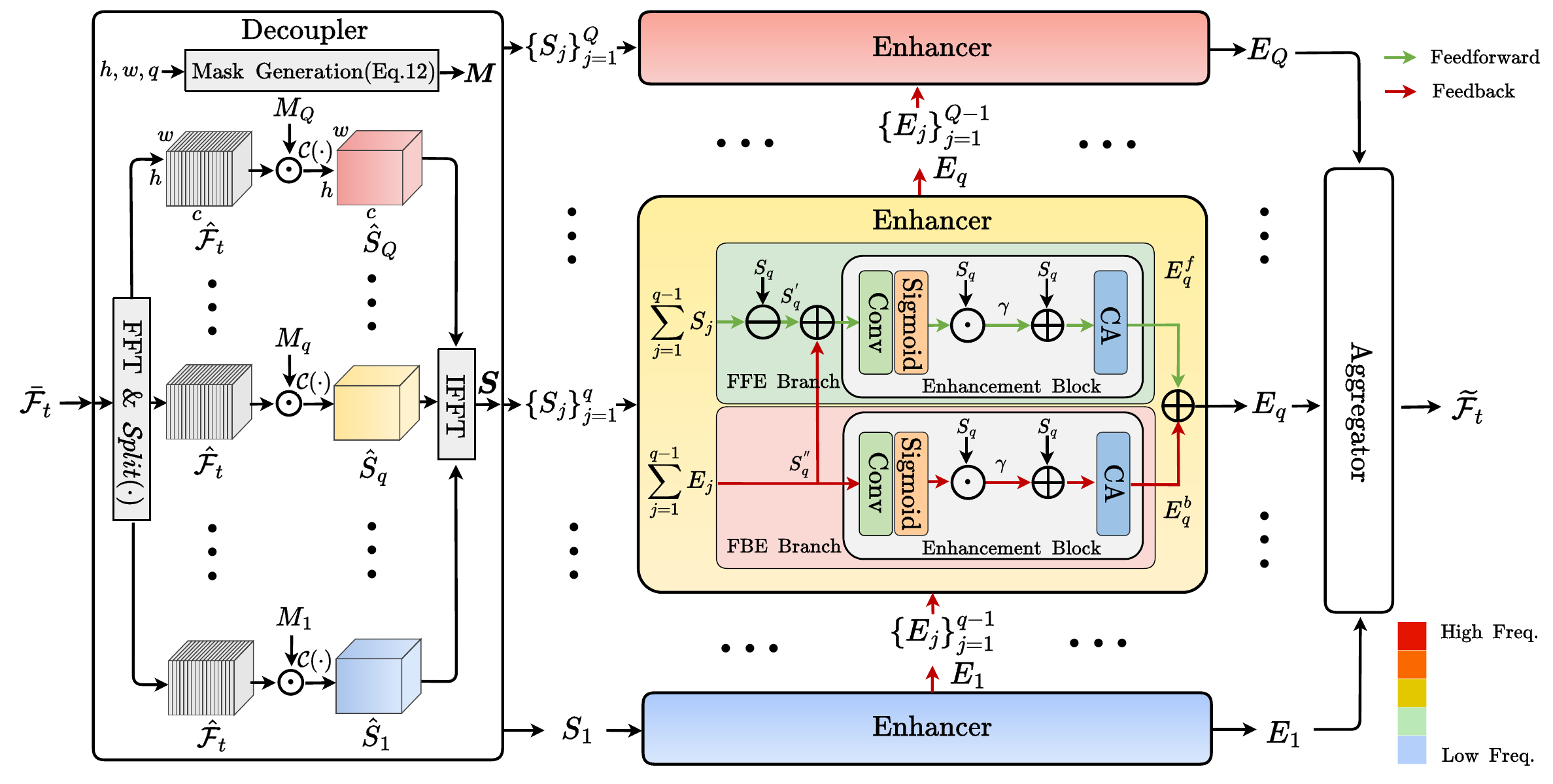}
\caption{The architecture of the multi-frequency feature refinement (MFFR) module.}
\label{vision_SFIF}
\end{figure*}

\subsubsection{Motion-Guided Adaptive Convolution Layer}

The existing optical flow-based alignment modules~\cite{liu2022temporal,chan2021basicvsr,xue2019video} or deformable convolution-based alignment modules~\cite{tian2020tdan,wang2019edvr} are typically associated with single motion and high complexity, restricting their adoption in practical applications. To achieve high performance and low complexity for feature alignment,
we utilize the estimated multiple motion offsets to independently guide the feature spatial sampling for each adaptive convolution in the MGAC layer based on predicted kernels. As shown in Fig.~\ref{fig_MGAA}, at the $n$-th adaptive convolution operation, $\mathbf{AC}\text{-}n$, the aligned feature $\bar{a}_n\in \mathbb{R}^{ h \times w \times c}$ is calculated as:
\begin{equation}
\bar{{a}}_n = \mathbf{AC}\text{-}n(\bar{{a}}_{n-1}, o_{n}, \mathbf{K}_{n})=  \mathbb{S}(\bar{{a}}_{n-1}, o_n) * \mathcal{K}^h_n * \mathcal{K}^v_n,
\end{equation}
where $n=1,\dots, N$, $\bar{a}_0$ = $\mathcal{F}_{t-3}$, $\bar{a}_N$ = $\bar{\mathcal{F}}_{t-2}$,  $\mathbb{S}(\cdot,\cdot)$ represents the spatial sampling operation and $*$ is the channel-wise convolution operator that performs convolutions in a spatially-adaptive manner. 
Our $\mathbf{AC}\text{-}n$ is composed of separable filters that ensure larger receptive fields at much lower computational costs than 2-dim convolutions. Based on this design, we apply $N$ adaptive convolutions to gradually sample the feature guided by motions and enable small-sized adaptive convolutions to be used for establishing large receptive fields. Benefit from these advantages, our MGAA module aims for better flexibility, higher performance, and lower complexity.

\subsection{Multi-Frequency Feature Refinement}

\begin{figure*}[!t]
\centering
\begin{minipage}[b]{0.094\linewidth}
\centering
\centerline{\epsfig{figure=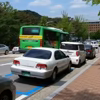,width=1.75cm}}
\footnotesize{LR Frame}  
\end{minipage}
\begin{minipage}[b]{0.094\linewidth}
\centering
\centerline{\epsfig{figure=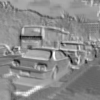,width=1.75cm}}
\footnotesize{$\bar{\mathcal{F}}_{t}$}  
\end{minipage}
\begin{minipage}[b]{0.094\linewidth}
\centering
\centerline{\epsfig{figure=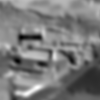,width=1.75cm}}
\footnotesize{$S_{1}$} 
\end{minipage}
\begin{minipage}[b]{0.094\linewidth}
\centering
\centerline{\epsfig{figure=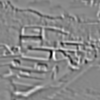,width=1.75cm}}
\footnotesize{$S_{2}$} 
\end{minipage}
\begin{minipage}[b]{0.094\linewidth}
\centering
\centerline{\epsfig{figure=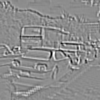,width=1.75cm}}
\footnotesize{$S_{3}$} 
\end{minipage}
\begin{minipage}[b]{0.094\linewidth}
\centering
\centerline{\epsfig{figure=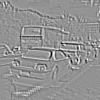,width=1.75cm}}
\footnotesize{$S_{4}$} 
\end{minipage}
\begin{minipage}[b]{0.094\linewidth}
\centering
\centerline{\epsfig{figure=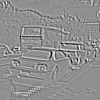,width=1.75cm}}
\footnotesize{$S_{5}$}  
\end{minipage}
\begin{minipage}[b]{0.094\linewidth}
\centering
\centerline{\epsfig{figure=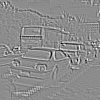,width=1.75cm}}
\footnotesize{$S_{6}$}  
\end{minipage}
\begin{minipage}[b]{0.094\linewidth}
\centering
\centerline{\epsfig{figure=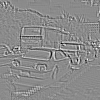,width=1.75cm}}
\footnotesize{$S_{7}$} 
\end{minipage}
\begin{minipage}[b]{0.094\linewidth}
\centering
\centerline{\epsfig{figure=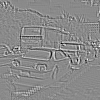,width=1.75cm}}
\footnotesize{$S_{8}$} 
\end{minipage}

\begin{minipage}[b]{0.094\linewidth}
\centering
\centerline{\epsfig{figure=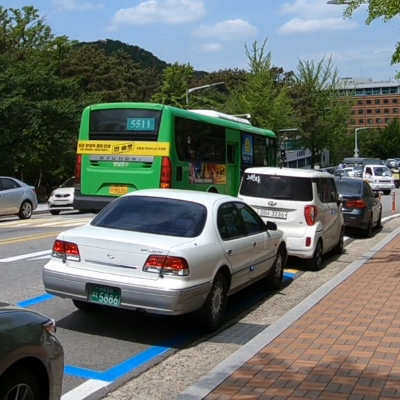,width=1.75cm}}
\footnotesize{GT}  
\end{minipage}
\begin{minipage}[b]{0.094\linewidth}
\centering
\centerline{\epsfig{figure=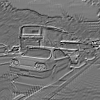,width=1.75cm}}
\footnotesize{$\mathcal{\widetilde{F}}_{t}$}  
\end{minipage}
\begin{minipage}[b]{0.094\linewidth}
\centering
\centerline{\epsfig{figure=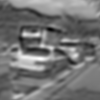,width=1.75cm}}
\footnotesize{$E_{1}$} 
\end{minipage}
\begin{minipage}[b]{0.094\linewidth}
\centering
\centerline{\epsfig{figure=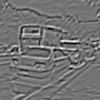,width=1.75cm}}
\footnotesize{$E_{2}$} 
\end{minipage}
\begin{minipage}[b]{0.094\linewidth}
\centering
\centerline{\epsfig{figure=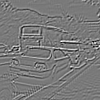,width=1.75cm}}
\footnotesize{$E_{3}$} 
\end{minipage}
\begin{minipage}[b]{0.094\linewidth}
\centering
\centerline{\epsfig{figure=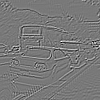,width=1.75cm}}
\footnotesize{$E_{4}$} 
\end{minipage}
\begin{minipage}[b]{0.094\linewidth}
\centering
\centerline{\epsfig{figure=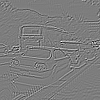,width=1.75cm}}
\footnotesize{$E_{5}$}  
\end{minipage}
\begin{minipage}[b]{0.094\linewidth}
\centering
\centerline{\epsfig{figure=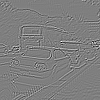,width=1.75cm}}
\footnotesize{$E_{6}$}  
\end{minipage}
\begin{minipage}[b]{0.094\linewidth}
\centering
\centerline{\epsfig{figure=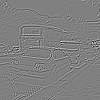,width=1.75cm}}
\footnotesize{$E_{7}$} 
\end{minipage}
\begin{minipage}[b]{0.094\linewidth}
\centering
\centerline{\epsfig{figure=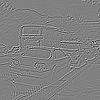,width=1.75cm}}
\footnotesize{$E_{8}$} 
\end{minipage}
\caption{{Visualization of input feature, output feature, decoupled features, enhanced features in the MFFR module.} }\label{visionMFFR}
\end{figure*}

Several works~\cite{li2023multi,dong2023dfvsr,zhu2023exploring} have been proposed to restore the high-frequency information in the frequency domain. However, these methods explore frequency information within the entire frequency range while they do not differentiate various frequency subbands, potentially leading to suboptimal results. Besides, different frequency subbands have different degrees of information loss, and higher frequency subbands usually lose more information in compression scenario. To fully explore high-frequency information and improve the SR performance, we designed a multi-frequency feature refinement (MFFR) module to differentiate frequency subbands and refine the input feature in different frequency subbands from low-frequency to high-frequency, as shown in Fig.~\ref{vision_SFIF}. It consists of \emph{Decoupler}, \emph{Enhancer}, and \emph{Aggregator} modules,  based on the decomposition-enhancement-aggregation strategy. 

Specifically, the \emph{Decoupler} module employs Gaussian band-pass filters to decompose the input feature $\bar{\mathcal{F}}_{t}\in \mathbb{R}^{h \times w \times c }$ into $Q$ features:
\begin{equation}
\textbf{\emph{S}} =  \left\{{S}_{j}\right\}_{j=1}^Q =   \mathrm{Decoupler}(\mathcal{\bar{F}}_{t}).
\end{equation}
The decomposed feature set $\textbf{\emph{S}}$ (or its subsets) is then fed into multiple \emph{Enhancer} modules to obtain the enhanced features $\emph{\textbf{E}}$=$\left\{{E}_{j} \right\}_{j=1}^Q$, ${E}_{j}\in\mathbb{R}^{h\times w \times c}$. Specifically, for the $q^{th}$ subband, the subset $\left\{S_j\right\}_{j=1}^{q-1}, S_j \in \mathbb{R}^{ h \times w \times c}$, and enhanced features $\left\{E_j\right\}_{j=1}^{q-1}$ for the lower subbands (if applicable) are input into the \emph{Enhancer} module with $S_q$ to obtain the enhanced feature $E_{q}$  at this subband level. This process is described by:
\begin{equation}
{E}_{q} = 
\begin{cases} 
\mathrm{Enhancer}({S}_{1}),  q = 1, \\
\mathrm{Enhancer}(\left\{S_j\right\}_{j=1}^q, \left\{E_j\right\}_{j=1}^{q-1}),  q = 2,\dots, Q.
\end{cases} 
\end{equation}
For the lowest subband, we additionally apply a mean filter on ${S}_{1}$ before inputting it into \emph{Enhancer}.

Finally, the \emph{Aggregator} module is employed to aggregate the enhanced features $\emph{\textbf{E}}$ and obtain the refined feature:
\begin{equation}
\mathcal{\widetilde{F}}_{t} = \mathrm{Aggregator}(\emph{\textbf{E}}).
\end{equation}

\subsubsection{Decoupler}
The workflow of the \emph{Decoupler} module is illustrated in Fig.~\ref{vision_SFIF}. To decompose the input feature $\mathcal{\bar{F}}_{t}$ into different frequency subbands, the input feature is first transformed to the frequency domain by FFT. We use the $\mathtt{torch.fft.fftn(\cdot)}$ function in Pytorch to generate the frequency feature with the same dimension as the input feature. The resulting frequency feature $\hat{\mathcal{F}}_{t} \in\mathbb{R}^{h \times w \times c}$ is then split along the channel dimension by $Split(\cdot)$ operation to obtain $c$ frequency channel features $\hat{\textbf{f}}=\left\{\hat{f}_{j}\right\}_{j=1}^c, \hat{f}_j \in\mathbb{R}^{h \times w \times 1}$. Sequentially, the \emph{Decoupler} module generates $Q$ Gaussian band-pass filter masks $\emph{\textbf{M}}$ = $\left\{M_{j}\right\}_{j=1}^Q, M_j \in\mathbb{R}^{h \times w \times 1}$. For each Gaussian band-pass filter mask $M_{j}$, its truncation frequency $d_j$ is calculated based on the width $h$ and height $w$ of the input feature:
\begin{equation}
d_j = \frac{j\sqrt{(\frac{h}{2})^2+(\frac{w}{2})^2}}{Q},
\end{equation}
and  $M_j $ is given by:
\begin{equation}
\begin{split}
M_j (u,v) = \exp \left(\frac{-\left[(u-h/2)^2+(v-w/2)^2\right]}{2 d_j^2}\right) \\
 -\sum_{l=1}^{j-1} \exp \left(\frac{-\left[(u-h/2)^2+(v-w/2)^2\right]}{2 d_l^2}\right).
\end{split} \label{BP_filter}
\end{equation}
The frequency channel features $\hat{\textbf{f}}=\left\{\hat{f}_{j}\right\}_{j=1}^c$ are multiplied by the band-pass filter mask $M_j$ and then concatenated to obtain the decomposed frequency feature $\hat{S}_{j}$:
\begin{equation}
\hat{S}_{j} = \mathcal{C}(M_j \odot \hat{{f}}_{1},..., M_j \odot \hat{{f}}_{t},...,M_j \odot \hat{{f}}_{c}),
\end{equation}
where $\odot$ is the element-wise multiply operation.

Finally, feature $\hat{S}_{j}\in \mathbb{R}^{ h \times w \times c}$ is transformed to the spatial domain through inverse FFT by $\mathtt{torch.fft.ifftn(\cdot)}$  function in Pytorch, producing the corresponding decomposed feature $S_j\in \mathbb{R}^{ h \times w \times c}$.

\subsubsection{Enhancer}

The video compression and downsampling operation mainly causes the loss of the high-frequency information in videos. To effectively restore the high-frequency information and emphasize the high-frequency subbands, we design the \emph{Enhancer} that progressively enhances the low-frequency subband and uses the enhanced low-frequency subbands to enhance the high-frequency subbands, which progressively optimizes the model to focus on the high-frequency, to beneficial for the recovery of high-frequency information. Decomposed feature set
$\left\{S_j\right\}_{j=1}^q$ and the enhanced feature set $\left\{E_j\right\}_{j=1}^{q-1}$ from the lower subbands are fed into the \emph{Enhancer} module to enhance the decomposed frequency feature ${S}_q$. The \emph{Enhancer} module consists of a feedforward enhancement (FFE) branch and a feedback enhancement (FBE) branch, both contain an enhancement block. For frequency subband $S_{q}$, the frequency subbands $\{{S_j}\}_{j=1}^{q-1}$ are the low-frequency subbands of $S_{q}$. As show in Fig.~\ref{vision_SFIF}, in the FFE branch, the input feature subset, $\left\{S_j\right\}_{j=1}^{q-1}$, is summed together, and subtracted by the decomposed feature $S_{q}$ to obtain a high-frequency feature $S^{'}_{q}$.
The enhanced feature set $\left\{E_j\right\}_{j=1}^{q-1}$ is summed together in the FBE branch, obtaining another high-frequency feature $S^{''}_{q}$. The sum of $S^{'}_{q}$ and $S^{''}_{q}$ is then input into the enhancement block, which consists of a 3$\times$3 convolution layer $\mathrm{Conv}$, a sigmoid activation function $\sigma$ and a channel attention ($\mathrm{CA}$)~\cite{zhang2018image}, to obtain the feedforward enhanced feature ${{E}}^{f}_{q}$.

In the FBE branch, 
$S^{''}_{q}$ is also processed by the enhancement block to obtain the feedback enhanced feature $E^{b}_{q}$, which will then be combined with ${{E}}^{f}_{q}$ to produce the final enhanced feature $E_{q}$. It is noted that when $q$ = 1 (corresponding to the lowest subband), we additionally apply a mean filter on $S_{q}$ which replaces $\left\{S_j\right\}_{j=1}^q$ as the input of the \emph{Enhancer} module and there is no FBE branch here.

\subsubsection{Aggregator}

To aggregate the enhanced frequency feature in each subband to generate the refined feature $\widetilde{{\mathcal{F}}}_{t}$, we use the following equation to sum them together before applying a channel attention ($\mathrm{CA}$)~\cite{zhang2018image} to strengthen the interaction between feature channels:
\begin{equation}
\mathcal{\widetilde{F}}_{t} = \mathrm{CA}(\sum_{j=1}^{Q} E_{j}),
\end{equation}
where $\mathrm{CA}$ consists of an average pooling layer, two convolution layers with $c$ channels, and a sigmoid activation function. 

Figure~\ref{visionMFFR} provides a visualization of the intermediate results generated in the MFFR module. The decomposed feature $S_i$ and the enhanced feature $E_i$ ($i=1,...,8$) in Fig.~\ref{visionMFFR} were extracted from a specific channel of the feature, i.e., 29th, and the channel features are saved as grayscale images for illustration. It can be observed that the resulting features at each stage exhibit the characteristics expected in the design - features corresponding to high-frequency subbands contain finer details, and vice versa.

\subsection{Reconstruction Module }

To generate an HR video from the refined feature $\mathcal{\widetilde{F}}_{t}$,  the scale-wise convolution block (SCB)~\cite{fan2020scale} with the residual-in-residual structure and a pixelshuffle layer are adopted to compose our reconstruction (REC) module. The REC module contains $R$ residual groups for information interaction.  Each residual group has three SCBs and a short skip connection. The output feature of $R$ residual groups is upsampled by a pixelshuffle layer to obtain the final HR residual frame ${I}^{\mathrm{Res}}_{t}$.

\subsection{Loss Functions}

The proposed model is optimized using the overall loss function $\mathcal{L}_{all}$ given below:
\begin{equation}
\mathcal{L}_{all} = \mathcal{L}_{spa} + \alpha  \mathcal{L}_{fc},
\end{equation}
where $\alpha$ is the weight factor, $ \mathcal{L}_{spa}$, $\mathcal{L}_{fc}$ are the spatial loss, frequency-aware contrastive loss, respectively, and their definition are provided below.

\subsubsection{Spatial Loss}

The Charbonnier loss function~\cite{lai2018fast} is adopted as our spatial loss function for supervising the generation of SR results in the spatial domain:
\begin{equation}
\mathcal{L}_{spa}=\sqrt{\left\|I_{t}^{\mathrm{HR}}-I_{t}^{\mathrm{SR}}\right\|^{2}+\epsilon^{2}},
\end{equation}
in which $I_{t}^{\mathrm{HR}}$ is the uncompressed HR frame and the penalty factor $\epsilon$ is set to $1 \times 10^{-4}$.

\begin{figure}[!t]
\centering
\includegraphics[width=0.495\textwidth]{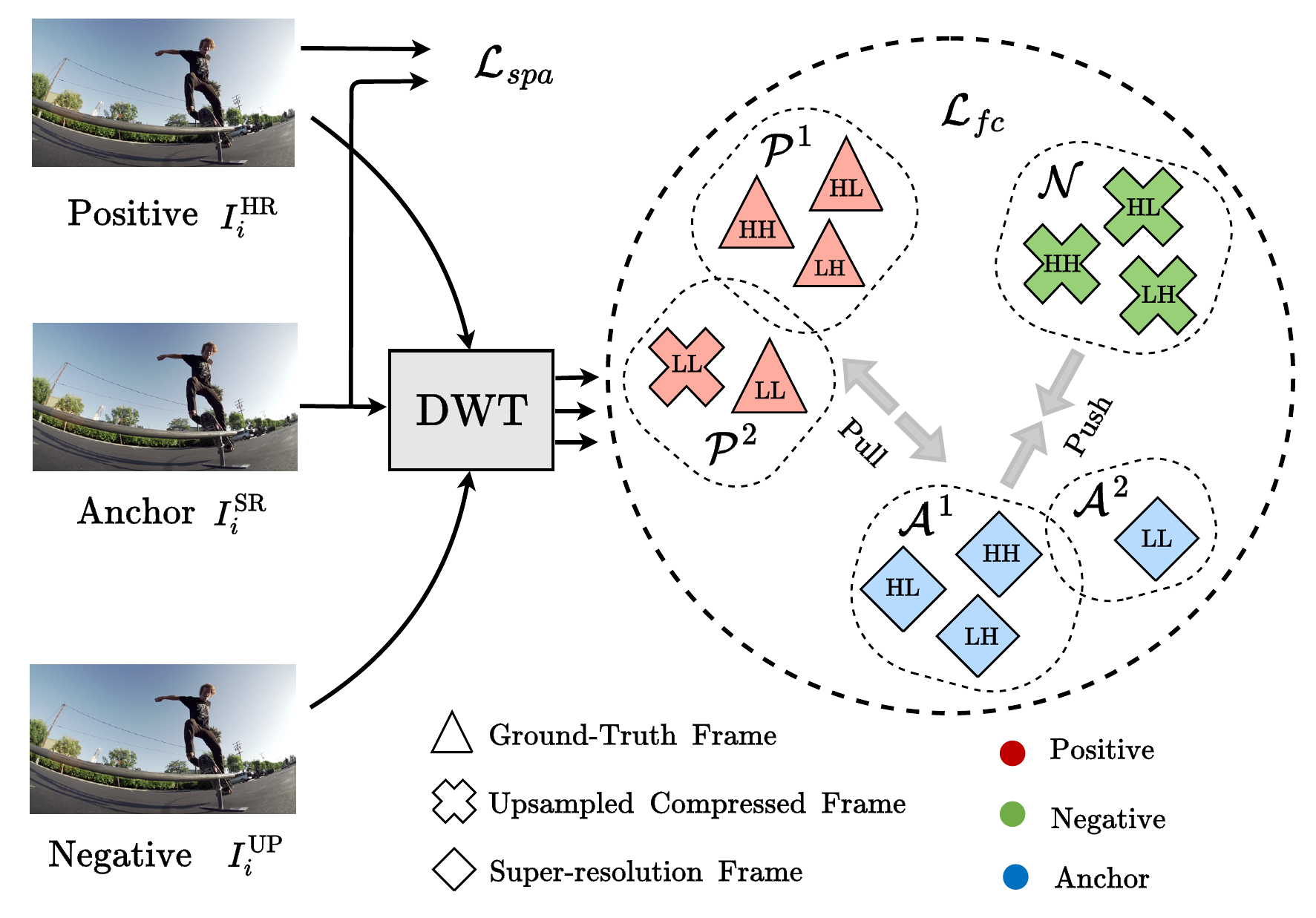}
\caption{The loss functions used for training the FCVSR model.}
\label{Fig_loss}
\end{figure}

\subsubsection{Frequency-aware Contrastive Loss}
Some wavelet-based loss functions~\cite{gao2024efficient,dong2023dfvsr} were proposed to only supervise the difference between subbands of the restored image and the HR image, which limits the exploration of more frequency information for supervision. In this part, the frequency-aware contrastive loss is designed based on the 2D discrete wavelet transform (DWT) and contrastive learning technique. Specifically,
given a training group with an bi-linearly upsampled compressed image $I_i^{\mathrm{UP}}$, the corresponding uncompressed HR image $I_i^{\mathrm{HR}}$  and the restored SR image $I_i^{\mathrm{SR}}$, 2D-DWT decomposes each of them into four frequency subbands: LL, HL, LH and HH. Two positive sets are defined as $\mathcal{P}^{1}_{i}$ = $\left\{I_i^{\mathrm{HR(HH)}}, I_i^{\mathrm{HR(HL)}}, I_i^{\mathrm{HR(LH)}} \right\}$ and $\mathcal{P}^{2}_{i}$ = $\left\{I_i^{\mathrm{HR(LL)}}, I_i^{\mathrm{UP(LL)}} \right\}$,  while one negative set is demoted as $\mathcal{N}_{i}$ = $\left\{I_i^{\mathrm{UP(HH)}}, I_i^{\mathrm{UP(HL)}}, I_i^{\mathrm{UP(LH)}}\right\}$. Two anchor sets $\mathcal{A}^{1}_{i}$ = $\left\{I_i^{\mathrm{SR(HH)}},  I_i^{\mathrm{SR(HL)}}, I_i^{\mathrm{SR(LH)}}\right\}$, $\mathcal{A}^{2}_{i}$ = $\left\{I_i^{\mathrm{SR(LL)}}\right\}$ are constructed. Based on these definitions, two frequency-aware contrastive losses for the $i$-th train group are:
\begin{equation}
\mathcal{L}^{1}_{i}=-\frac{1}{G^{1}_p} \sum_{l=1}^{G^{1}_p}\log \frac{\exp (s(a^{1}_l, p^{1}_l) / \tau) }{\exp (s(a^{1}_l, p^{1}_l) / \tau)+\sum_{k=1}^{G_g} \exp (s(a^{1}_l, g_k) / \tau)},
\end{equation}
\begin{equation}
\mathcal{L}^{2}_{i}=- \frac{1}{G^{2}_p} \sum_{l=1}^{G^{2}_p}\log \frac{\exp (s(a^{2}, p^{2}_l) / \tau) }{\exp (s(a^{2}, p^{2}_l) / \tau)+\sum_{k=1}^{G_g} \exp (s(a^{2}, g_k) / \tau)},
\end{equation}
where $G^{1}_{p}$, $G^{2}_{p}$ and $G_{g}$ are the number of sets $\mathcal{P}^{1}$, $\mathcal{P}^{2}$ and $\mathcal{N}$, $\tau$ is the temperature parameter and $s(\cdot ,\cdot)$ is the similarity function.
$a$, $p$, and $g$ represent the anchor, positive, and negative samples, respectively.

The total frequency-aware contrastive loss is defined as:
\begin{equation}
\mathcal{L}_{fc} = \frac{1}{M_s} \sum_{i=1}^{M_s} ( \mathcal{L}_i^{1}+\mathcal{L}_i^{2}),
\end{equation}
where $M_s$ is the number of samples.

\begin{table*}[!t]
\caption{Quantitative comparison in terms of  PSNR (dB), SSIM and VMAF on three public testing datasets under LDB configuration. The FLOPs is calculated on LR video frame with 64 $\times$ 64 resolution and FPS is calculated on the REDS4 dataset. The best and the second-best results are highlighted and underlined.}\label{tab_PSNR}
\setlength{\tabcolsep}{1.0mm}
\renewcommand\arraystretch{1.1}
\fontsize{7}{9}\selectfont
\centering
\begin{tabular}{c|r|rrr|c|c|c|c}
\toprule
{\multirow{2}{*}{{Datasets} }} &  {\multirow{2}{*}{ {Methods} }} & \multicolumn{1}{c}{Param.$\downarrow$} & \multicolumn{1}{c}{FLOPs$\downarrow$} & \multicolumn{1}{c|}{FPS$\uparrow$}  & \multicolumn{1}{c|}{{QP = 22}} & \multicolumn{1}{c|}{{QP = 27}} & \multicolumn{1}{c|}{{QP = 32}}  & \multicolumn{1}{c}{{QP = 37}}  \\
&   & \multicolumn{1}{c}{(M)} & \multicolumn{1}{c}{(G)}  & \multicolumn{1}{c|}{(1/s)} & {PSNR}$\uparrow$ / {SSIM}$\uparrow$ / VMAF$\uparrow$  & {PSNR}$\uparrow$ / {SSIM}$\uparrow$ / VMAF$\uparrow$ & {PSNR}$\uparrow$ / {SSIM}$\uparrow$ / VMAF$\uparrow$ & {PSNR}$\uparrow$ / {SSIM}$\uparrow$ / VMAF$\uparrow$ \\
\midrule
&  Bicubic & -   & -  & -    & {29.39 / 0.8144 / 50.67}   &  {28.71 / 0.7949 / 40.78}  &  {27.74 / 0.7698 / 29.26}  & {26.42 / 0.7388 / 16.80}   \\
&  EDVR-L~\cite{wang2019edvr} & 20.69   & 354.07  & 2.02    & 31.76 / 0.8629 / 68.23   & 30.58 / {0.8377} / 56.39  & {29.07} / 0.8045 / 41.72  & 27.38 / 0.7670 / 25.53   \\
&  {BasicVSR}~\cite{chan2021basicvsr} & \underline{6.30}  & 367.72  & 0.85    & 31.80 / 0.8631  / 76.44  & 30.46 / 0.8349 / 65.14   & 29.05 / 0.8031 / 47.06   & 27.33 / 0.7661 / 29.59    \\
\multirow [c]{2}{*}{\rotatebox{90}{CVCP10~\cite{chen2021compressed} }}
&  {IconVSR}~\cite{chan2021basicvsr} & 8.70  & 576.45  & 0.51    & 31.86 / 0.8637 / 77.94  & 30.48 / 0.8354 /  64.69    & 29.10 / 0.8043 / 47.77   & 27.40 / 0.7678 / 30.05   \\
&  {BasicVSR++}~\cite{chan2022basicvsr++} & 7.32  & 395.69  & 0.74   & {31.89} / {0.8647} / 77.55    & {30.66} / {0.8388} / 66.43    & {29.13} / {0.8058} / 50.08    & {27.43} / {0.7682} / 34.11     \\
&  {FTVSR++}~\cite{qiu2023learning} & 10.80   & 1148.85  & 0.27  &  {31.92 / 0.8656 / 78.52}    & {30.69 / 0.8393 / 66.89}  & {29.14 / 0.8063 / 51.96}    &  {27.44} / 0.7697 / 35.06  \\
&  {MIA-VSR}~\cite{zhou2024video} & {16.59}   & {1402.32}  & {0.23}  &  31.92 / 0.8660 / 78.58    & 30.68 / 0.8394 / 66.91  & 29.15 / 0.8068 / 51.85    &  \underline{27.45} / 0.7694 / 35.16  \\
&  {IA-RT}~\cite{xu2024enhancing} & {13.41}   &  {1939.50}  &  {0.35}  &  \textbf{31.95} / \underline{0.8668} / \textbf{78.71}    & \textbf{30.72 / 0.8406 / 67.01}  & \underline{29.17 / 0.8074 / 52.00}    &  \textbf{27.46 / 0.7706} / \underline{35.46}  \\
\cmidrule{2-9}
&  {FCVSR-S (ours)}  & \textbf{3.70}  & \textbf{68.82}  & \textbf{5.28}   & 31.86 / 0.8650 / 78.27 &  30.64 / 0.8388 / 65.96  & 29.10 / 0.8058  / 51.39  & {27.44 / 0.7700 / 35.07}   \\
&  {FCVSR (ours)}  & 8.81  & \underline{165.36}  & \underline{2.39}    & \underline{31.94} / \textbf{0.8669}  / \underline{78.69}  & \underline{30.70 / 0.8403 / 66.97}  &  \textbf{29.18 / 0.8077 / 52.03}   &  \textbf{27.46} / \underline{0.7704} / \textbf{35.63} \\
\midrule
&  {Bicubic} & -   & -  & -    & {26.94 / 0.7271 / 50.98}   &  {26.39 / 0.7027 / 40.07}  & {25.58 / 0.6726 / 28.32}  &  {24.47 / 0.6377 / 17.60}   \\
&  EDVR-L~\cite{wang2019edvr} & 20.69   & 354.07  & 2.02   & 29.05 / 0.7991 / 81.60 & 27.60 / 0.7470 / 59.90   &  26.40 / 0.7072 / 46.31   &   24.87 / 0.6585 / 28.80  \\
&  {BasicVSR}~\cite{chan2021basicvsr} & \underline{6.30}  & 367.72  & 0.85  & 29.13 / 0.8005 / 81.13  & 27.62 / 0.7512 / 63.49 & 26.43 / 0.7079 / 46.82   & 24.99 / 0.6603 / 29.49 \\
\multirow [c]{2}{*}{\rotatebox{90}{REDS4~\cite{nah2019ntire}} }
&  IconVSR~\cite{chan2021basicvsr}  & 8.70  & 576.45  & 0.51   & 29.17 / 0.8009 / 81.52  & 27.73 / 0.7519 / 62.91  & 26.45 / 0.7090 / 47.48  & 24.99 / 0.6609 / 29.73 \\
&   {BasicVSR++}~\cite{chan2022basicvsr++} & 7.32  & 395.69  & 0.74   &   29.23 / 0.8036 / 81.83  & 27.79 / 0.7543 / 63.63  & 26.50 / 0.7098 / 47.78   &  25.05 / 0.6620 / 31.25 \\
& {FTVSR++}~\cite{qiu2023learning} & 10.80   & 1148.85  & 0.27  &  {29.26 / 0.8029 / 81.58} &  {27.81 / 0.7564 / 65.22} &  {26.53 / 0.7106 / 48.57}  &  {25.09 / 0.6625 / 31.81}  \\
&  {MIA-VSR}~\cite{zhou2024video} & {16.59}   & {1402.32}  & {0.23}   &  29.27 / 0.8032 / 81.67    & \underline{27.85} / 0.7578 / 65.45   &  26.58 / 0.7137 / 48.36 & 25.14 / 0.6652 / 31.94     \\
&  {IA-RT}~\cite{xu2024enhancing} & {13.41}   &  {1939.50}  &  {0.35}   &  \textbf{29.31 / 0.8041 / 81.90}    & \textbf{27.92} / \textbf{0.7593} / \underline{65.58}  & \underline{26.62 / 0.7155 / 48.50}    &  \underline{25.16 / 0.6678 / 32.01}  \\
\cmidrule{2-9}
&  {FCVSR-S (ours)}  & \textbf{3.70}  & \textbf{68.82}  & \textbf{5.28}   &   29.14 / 0.8002 / 81.18   &  27.66 / 0.7505 / 63.14   &  26.42 / 0.7089 / 47.75   &  24.93 / 0.6611 / 31.56 \\
&  {FCVSR (ours)}  & 8.81  & \underline{165.36}  & \underline{2.39}   &  \underline{29.28 / 0.8039 / 81.87}   & \textbf{27.92} / \underline{0.7591}  / \textbf{65.63}   & \textbf{26.64 / 0.7161  / 48.59}   &   \textbf{25.20 / 0.6694 / 32.05} \\
\midrule
&  {Bicubic}  & -   & -  & -    & {23.53 / 0.6097 / 31.61}   &  {23.18 / 0.5836 / 22.86}  &  {22.60 / 0.5454 / 11.52}  & {21.75 / 0.4954 / 2.48 }   \\
&  EDVR-L~\cite{wang2019edvr}  & 20.69   & 354.07  & 2.02  & 25.27 / 0.7135 / 66.57  &   24.31 / 0.6586 / 52.82   & 23.29 / 0.5958 / 34.74  & 22.09 / 0.5284 / 20.43  \\
&  BasicVSR~\cite{chan2021basicvsr}  & \underline{6.30}  & 367.72  & 0.85  & 25.30 / 0.7155 / 67.23  &  24.36 / 0.6610 / 52.69    & 23.34 / 0.5989 / 35.51  &   22.15 / 0.5314 / 20.52 \\
\multirow [c]{2}{*}{\rotatebox{90}{Vid4~\cite{xue2019video}  } }
&  {IconVSR}~\cite{chan2021basicvsr}  & 8.70  & 576.45  & 0.51 & 25.46 / 0.7225 / 68.77   & 24.41 / 0.6638 / 52.88  &  23.36 / 0.5993 / 35.53 &   22.16 / 0.5305 / 20.41  \\
&  {BasicVSR++}~\cite{chan2022basicvsr++}& 7.32  & 395.69  & 0.74   & 25.55 / 0.7270 / 70.35   &  24.43 / 0.6639 / 53.93  & 23.37 / 0.5976 / 35.30  & 22.18 / 0.5326 / 20.60  \\
&  {FTVSR++}~\cite{qiu2023learning} & 10.80   & 1148.85  & 0.27 & {25.58 / 0.7278 / 70.68}  & {24.44 / 0.6657} / 53.53   &   23.39 / {0.6024} / 36.16 &  {22.20} / 0.5338 / 20.90 \\
&  {MIA-VSR}~\cite{zhou2024video} & {16.59}   & {1402.32}  & {0.23}   &  25.60 / 0.7288 / 70.96    & \underline{24.50} / 0.6696 / 54.07  & 23.42 / 0.6039 / 36.72    &  22.23 / 0.5346 / 21.36  \\
&  {IA-RT}~\cite{xu2024enhancing} & {13.41}   &  {1939.50}  &  {0.35}    &  \textbf{25.64 / 0.7310} / \underline{71.26}    & \textbf{24.58} / \underline{0.6705 / 54.65}  & \underline{23.46 / 0.6049 / 37.11}    &  \textbf{22.27} / \underline{0.5359 / 21.52}  \\
\cmidrule{2-9}
&  {FCVSR-S (ours)}  & \textbf{3.70}  & \textbf{68.82}  & \textbf{5.28}   & 25.35 / 0.7194 / 68.36  & 24.43 / 0.6647 / 53.50    & {23.40} / 0.6021 / {36.25}    &   22.19 / {0.5340 / 21.08} \\
&  {FCVSR (ours)}  & 8.81  & \underline{165.36}  & \underline{2.39}   & \underline{25.61 / 0.7307} / \textbf{71.50}  & \textbf{24.58 / 0.6707 / 54.79} &  \textbf{23.47 / 0.6052 / 37.20}  & \underline{22.25} / \textbf{0.5366 / 21.60} \\
\bottomrule	
\end{tabular}
\end{table*}

\section{Experiment and Results} \label{EXP}

\subsection{Implementation Details}
In this work, an FCVSR model and its lightweight model, i.e., FCVSR-S, are proposed for the compressed VSR task. The FCVSR model employs the following hyper parameters and configurations: the number of adaptive convolutions in the MGAA module is set as $N$ = 6; the decomposition number in the MFFR module is set as $Q$ = 8; the number of residual groups in the REC module is set as $R$ = 10. The FCVSR-S model is associated with lower computational complexity. Its number of adaptive convolutions in the MGAA module is set as $N$ = 4, the decomposition number of the decoupler is set as $Q$ = 4 in the MFFR module, and the number of residual groups is set as $R$ = 3 in the REC module.  The proportional factor is set as $\gamma$ = 0.2 in enhancement blocks. 
These two models all take 7 frames as the model input and use the overall loss function to train them. The weight factor of the overall loss function $\alpha$ is set to 1. The $L_{1}$ distance is adopted as the similarity function $s(\cdot)$ and the temperature parameter $\tau$ is 1 in $\mathcal{L}_{fc}$. The compressed frames are cropped into 128$\times$128 patches and the batch size is set to 8. Random rotation and reflection operations are adopted to increase the diversity of training data. The proposed models are implemented based on PyTorch and trained by Adam~\cite{kingma2014adam}. The learning rate is initially set to 2$\times$$10^{-4}$ and gradually halves at 2K, 8K and 12K epochs. The total number of epochs is 30K. All experiments are conducted on PCs with RTX-3090 GPUs and Intel Xeon(R) Gold 5218 CPUs.

\subsection{Experimental Setup}

Following the common practice in the previous works~\cite{chan2021basicvsr,chan2022basicvsr++,qiu2023learning}, our models are trained separately on three public training datasets, CVCP~\cite{chen2021compressed}, REDS~\cite{nah2019ntire} and Vimeo-90K~\cite{xue2019video}, and evaluated their corresponding test sets, CVCP10~\cite{chen2021compressed}, REDS4~\cite{nah2019ntire}, and Vid4~\cite{xue2019video} respectively. 
The downsampled LR videos are generated using a Bicubic filter with a factor of 4. All training and test compressed videos are created using the downsampling-then-encoding procedure and compressed by HEVC HM 16.20~\cite{peixoto2013h} under the Low Delay B mode with four different QP values: 22, 27, 32 and 37. Due to the CVCP dataset offers the Y channel of
YCbCr space videos, the channel of image $c_I$ is 1. $c_I$ is 3 for REDS and Vimeo-90K.

The peak signal-to-noise ratio (PSNR), structural similarity index (SSIM)~\cite{wang2004image}, and video multi-method assessment fusion (VMAF)~\cite{li2018vmaf} are adopted as evaluation metrics for the quantitative benchmark. PSNR and SSIM were widely used to evaluate the quality of videos, while VMAF was proposed by Netflix to evaluate the perceptual quality of videos. We also measured the model complexity in terms of the floating point operations (FLOPs), inference speed (FPS), and the number of model parameters.

\begin{figure*}[!t]
\centering
\begin{minipage}[b]{0.17\linewidth}
\centering
\centerline{\epsfig{figure=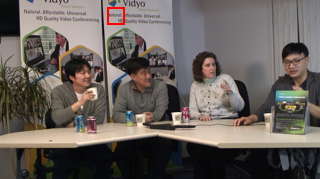,width=3.12cm}}
\footnotesize{\emph{CVCP10\_FourPeople\_011} \\ (QP=22)} 
\end{minipage}
\begin{minipage}[b]{0.095\linewidth}
\centering
\centerline{\epsfig{figure=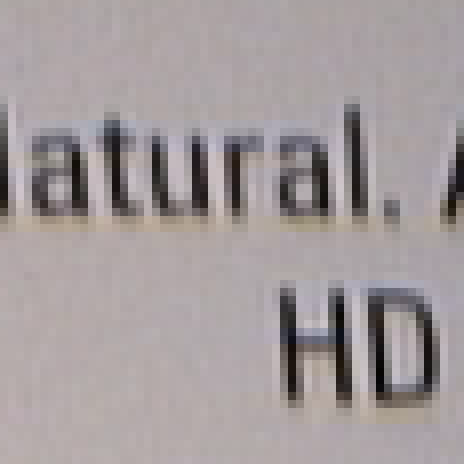,width=1.75cm}}
\footnotesize{GT \\ \  } 
\end{minipage}
\begin{minipage}[b]{0.095\linewidth}
\centering
\centerline{\epsfig{figure=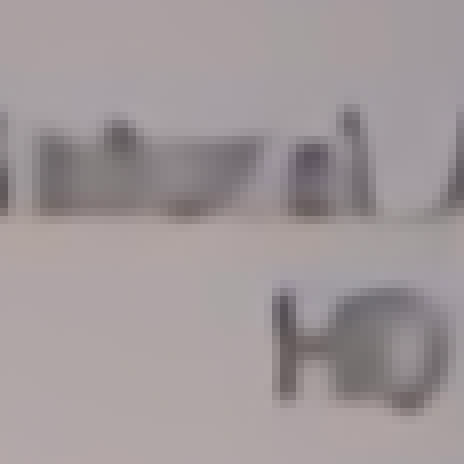,width=1.75cm}}
\footnotesize{  IconVSR \\ \  }
\end{minipage}
\begin{minipage}[b]{0.095\linewidth}
\centering
\centerline{\epsfig{figure=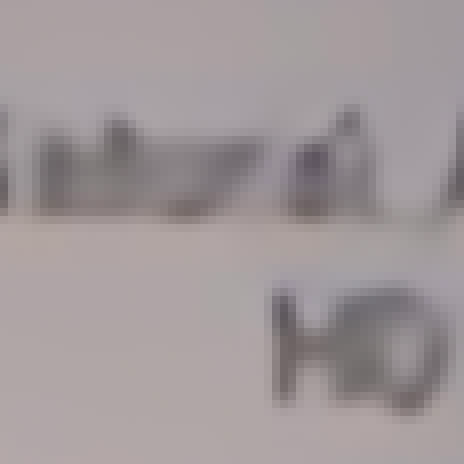,width=1.75cm}}
\footnotesize{ BasicVSR++ \\ \  }
\end{minipage}
\begin{minipage}[b]{0.095\linewidth}
\centering
\centerline{\epsfig{figure=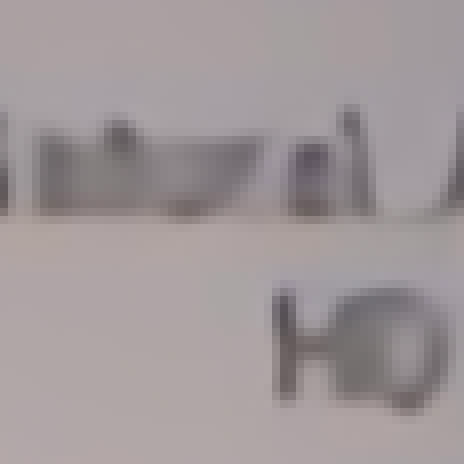,width=1.75cm}}
\footnotesize{  FTVSR++ \\ \  }
\end{minipage}
\begin{minipage}[b]{0.095\linewidth}
\centering
\centerline{\epsfig{figure=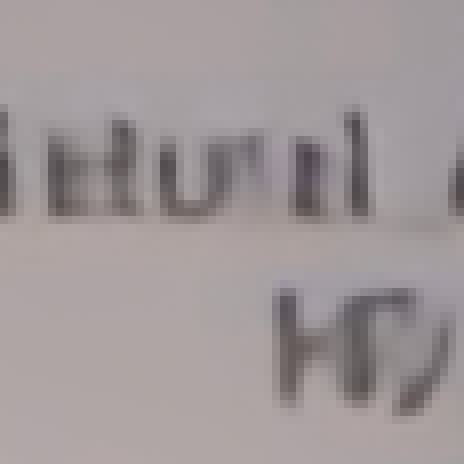,width=1.75cm}}
\footnotesize{ { MIA-VSR} \\ \  }
\end{minipage}
\begin{minipage}[b]{0.095\linewidth}
\centering
\centerline{\epsfig{figure=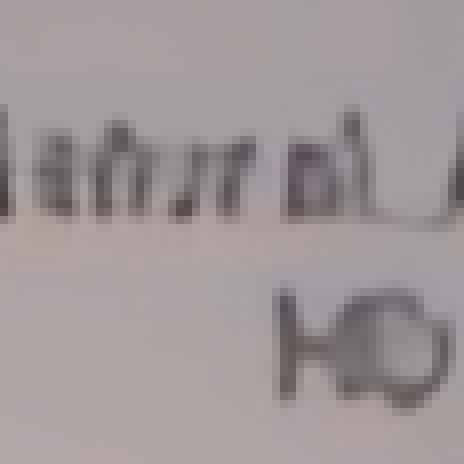,width=1.75cm}}
\footnotesize{ { IA-RT }\\ \  }
\end{minipage}
\begin{minipage}[b]{0.095\linewidth}
\centering
\centerline{\epsfig{figure=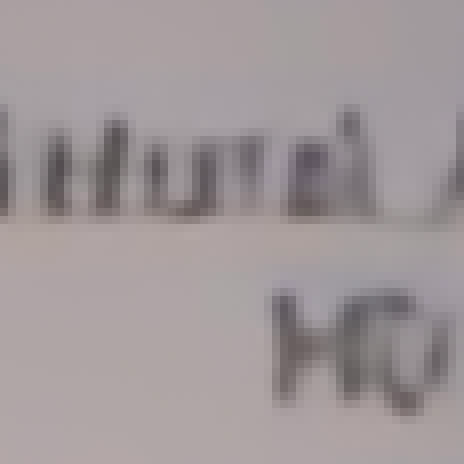,width=1.75cm}}
\footnotesize{  {FCVSR-S} \\ \  }
\end{minipage}
\begin{minipage}[b]{0.095\linewidth}
\centering
\centerline{\epsfig{figure=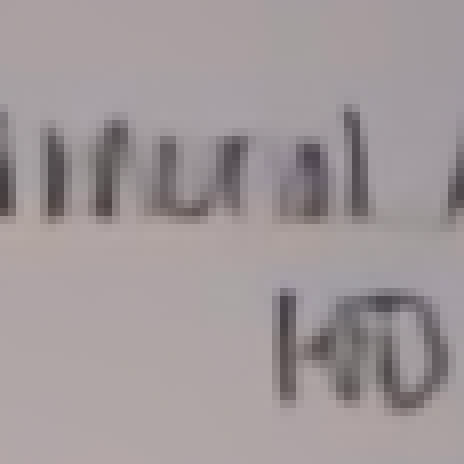,width=1.75cm}}
\footnotesize{ {FCVSR}  \\ \  }
\end{minipage}

\begin{minipage}[b]{0.17\linewidth}
\centering
\centerline{\epsfig{figure=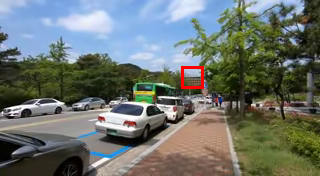,width=3.12cm}}
\footnotesize{\emph{REDS4\_000\_024} (QP=27)} 
\end{minipage}
\begin{minipage}[b]{0.095\linewidth}
\centering
\centerline{\epsfig{figure=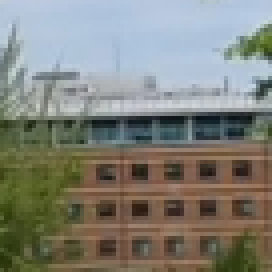,width=1.75cm}}
\footnotesize{GT } 
\end{minipage}
\begin{minipage}[b]{0.095\linewidth}
\centering
\centerline{\epsfig{figure=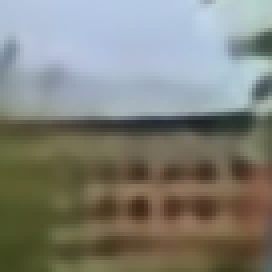,width=1.75cm}}
\footnotesize{  IconVSR  }
\end{minipage}
\begin{minipage}[b]{0.095\linewidth}
\centering
\centerline{\epsfig{figure=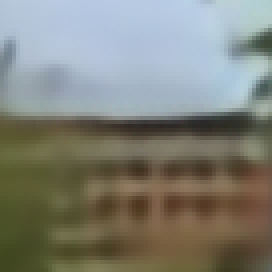,width=1.75cm}}
\footnotesize{  BasicVSR++  }
\end{minipage}
\begin{minipage}[b]{0.095\linewidth}
\centering
\centerline{\epsfig{figure=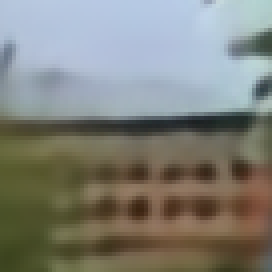,width=1.75cm}}
\footnotesize{  FTVSR++ }
\end{minipage}
\begin{minipage}[b]{0.095\linewidth}
\centering
\centerline{\epsfig{figure=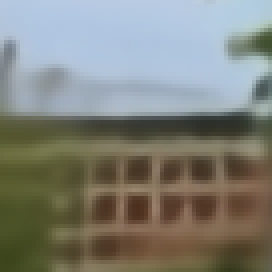,width=1.75cm}}
\footnotesize{ { MIA-VSR}  }
\end{minipage}
\begin{minipage}[b]{0.095\linewidth}
\centering
\centerline{\epsfig{figure=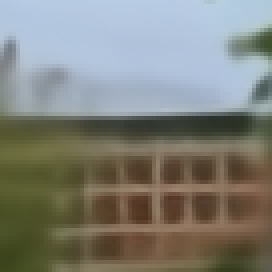,width=1.75cm}}
\footnotesize{{ IA-RT } }
\end{minipage}
\begin{minipage}[b]{0.095\linewidth}
\centering
\centerline{\epsfig{figure=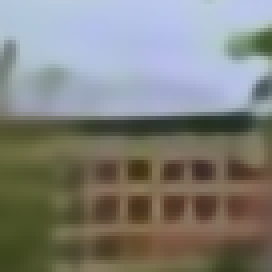,width=1.75cm}}
\footnotesize{  {FCVSR-S} }
\end{minipage}
\begin{minipage}[b]{0.095\linewidth}
\centering
\centerline{\epsfig{figure=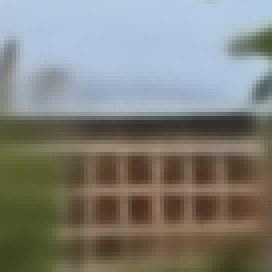,width=1.75cm}}
\footnotesize{ {FCVSR} }
\end{minipage}

\begin{minipage}[b]{0.17\linewidth}
\centering
\centerline{\epsfig{figure=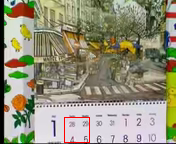,width=3.050cm,height=2.52cm}}
\footnotesize{\emph{Vid4\_Calendar\_020} (QP=32)  }  
\end{minipage}
\begin{minipage}[b]{0.095\linewidth}
\centering
\centerline{\epsfig{figure=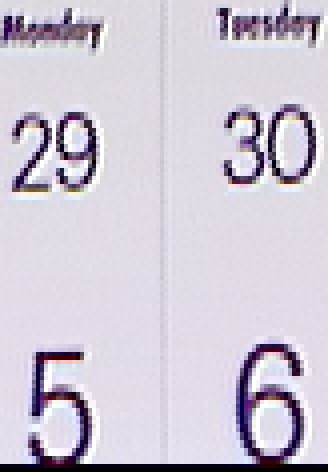,width=1.75cm}}
\footnotesize{GT \\ \  } 
\end{minipage}
\begin{minipage}[b]{0.095\linewidth}
\centering
\centerline{\epsfig{figure=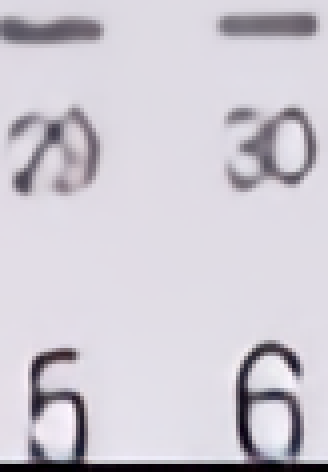,width=1.75cm}}
\footnotesize{  IconVSR  \\ \ }
\end{minipage}
\begin{minipage}[b]{0.095\linewidth}
\centering
\centerline{\epsfig{figure=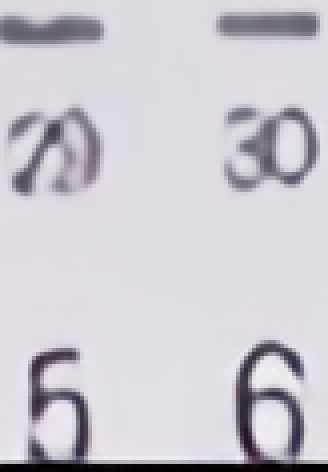,width=1.75cm}}
\footnotesize{ BasicVSR++  \\ \ }
\end{minipage}
\begin{minipage}[b]{0.095\linewidth}
\centering
\centerline{\epsfig{figure=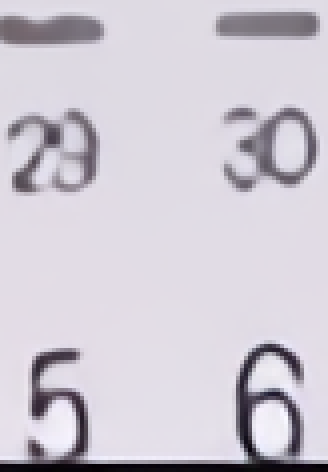,width=1.75cm}}
\footnotesize{ FTVSR++ \\ \   }
\end{minipage}
\begin{minipage}[b]{0.095\linewidth}
\centering
\centerline{\epsfig{figure=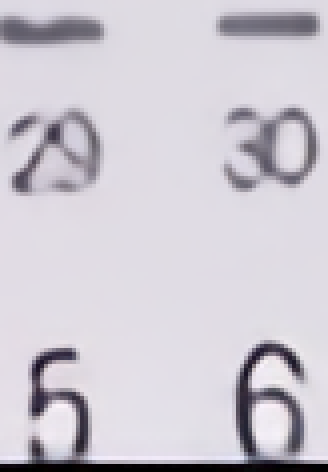,width=1.75cm}}
\footnotesize{  { MIA-VSR}  \\ \  }
\end{minipage}
\begin{minipage}[b]{0.095\linewidth}
\centering
\centerline{\epsfig{figure=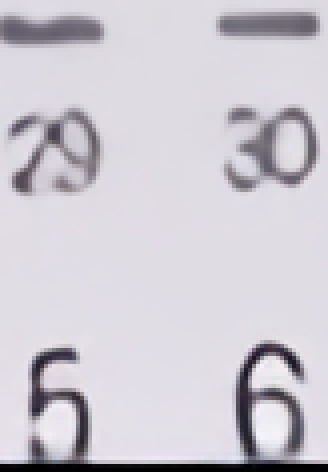,width=1.75cm}}
\footnotesize{{ IA-RT } \\ \  }
\end{minipage}
\begin{minipage}[b]{0.095\linewidth}
\centering
\centerline{\epsfig{figure=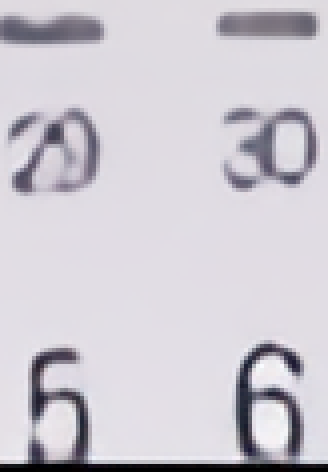,width=1.75cm}}
\footnotesize{ {FCVSR-S} \\ \  }
\end{minipage}
\begin{minipage}[b]{0.095\linewidth}
\centering
\centerline{\epsfig{figure=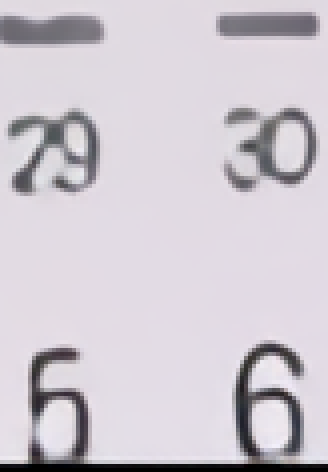,width=1.75cm}}
\footnotesize{ {FCVSR} \\ \  }
\end{minipage}

\begin{minipage}[b]{0.17\linewidth}
\centering
\centerline{\epsfig{figure=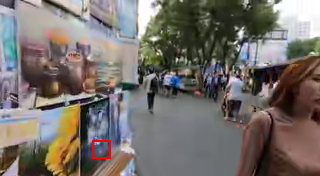,width=3.12cm}}
\footnotesize{\emph{REDS4\_020\_069} (QP=37)}  
\end{minipage}
\begin{minipage}[b]{0.095\linewidth}
\centering
\centerline{\epsfig{figure=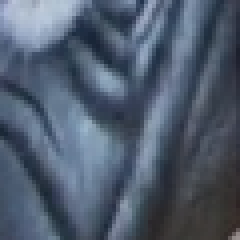,width=1.75cm}}
\footnotesize{GT  }
\end{minipage}
\begin{minipage}[b]{0.095\linewidth}
\centering
\centerline{\epsfig{figure=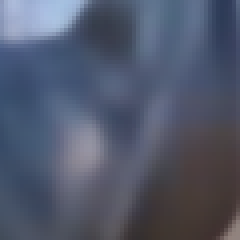,width=1.75cm}}
\footnotesize{  IconVSR }
\end{minipage}
\begin{minipage}[b]{0.095\linewidth}
\centering
\centerline{\epsfig{figure=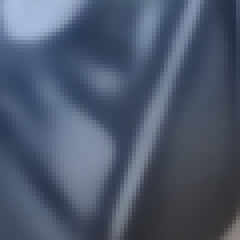,width=1.75cm}}
\footnotesize{ BasicVSR++}
\end{minipage}
\begin{minipage}[b]{0.095\linewidth}
\centering
\centerline{\epsfig{figure=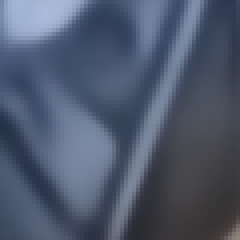,width=1.75cm}}
\footnotesize{ FTVSR++ }
\end{minipage}
\begin{minipage}[b]{0.095\linewidth}
\centering
\centerline{\epsfig{figure=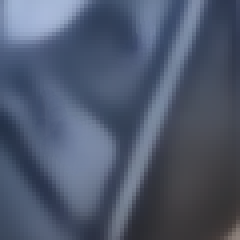,width=1.75cm}}
\footnotesize{  { MIA-VSR} }
\end{minipage}
\begin{minipage}[b]{0.095\linewidth}
\centering
\centerline{\epsfig{figure=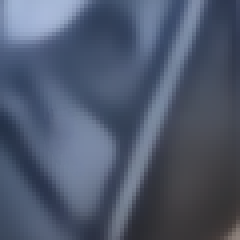,width=1.75cm}}
\footnotesize{ { IA-RT }}
\end{minipage}
\begin{minipage}[b]{0.095\linewidth}
\centering
\centerline{\epsfig{figure=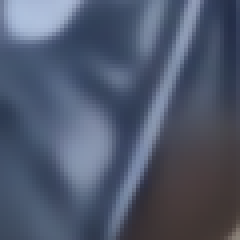,width=1.75cm}}
\footnotesize{ {FCVSR-S}}
\end{minipage}
\begin{minipage}[b]{0.095\linewidth}
\centering
\centerline{\epsfig{figure=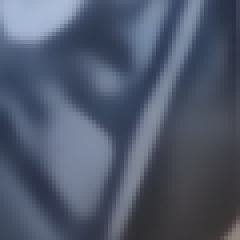,width=1.75cm}}
\footnotesize{ {FCVSR} }
\end{minipage}
\caption{{Visual comparison results between FCVSR models and five benchmark methods.}}
\label{vision_RST}
\end{figure*}

Seven state-of-the-art methods including EDVR-L~\cite{wang2019edvr}, BasicVSR~\cite{chan2021basicvsr}, IconVSR~\cite{chan2021basicvsr},  BasicVSR++~\cite{chan2022basicvsr++}, FTVSR++~\cite{qiu2023learning}, MIA-VSR~\cite{zhou2024video} and IA-RT~\cite{xu2024enhancing}  are benchmarked against the proposed models.  To ensure a fair comparison, we retrained EDVR-L~\cite{wang2019edvr}, BasicVSR~\cite{chan2021basicvsr},  IconVSR~\cite{chan2021basicvsr}, BasicVSR++~\cite{chan2022basicvsr++}, FTVSR++~\cite{qiu2023learning},  MIA-VSR~\cite{zhou2024video}, and IA-RT~\cite{xu2024enhancing} following the same training-evaluation procedure as the FCVSR model, using their publicly released source code. 

\begin{table*}[!t]
\caption{Quantitative Comparison in terms of PSNR (dB) and SSIM on the compressed videos of REDS4 and Vid4 datasets. The best and the second-best results are highlighted and underlined.} \label{Tab_CRF}
\setlength{\tabcolsep}{4.3mm}
\fontsize{7}{9}\selectfont
\centering
\begin{tabular}{r|c|c|c|c|c|c}
\toprule 
\multirow{4}{*}{Methods} & \multicolumn{3}{c|}{REDS4~\cite{nah2019ntire}} & \multicolumn{3}{c}{Vid4~\cite{xue2019video}} \\
\cmidrule{2-7}
& CRF = 15 & CRF = 25 & CRF = 35    & CRF = 15 & CRF = 25 & CRF = 35 \\
& PSNR$\uparrow$ / SSIM$\uparrow$  & PSNR$\uparrow$ / SSIM$\uparrow$ & PSNR$\uparrow$ / SSIM$\uparrow$ & PSNR$\uparrow$ / SSIM$\uparrow$ & PSNR$\uparrow$ / SSIM$\uparrow$  & PSNR$\uparrow$ / SSIM$\uparrow$  \\ 
\midrule 
DUF~\cite{jo2018deep}      & 25.61 / 0.775 & 24.19 / 0.692 & 22.17 / 0.588 & 24.40 / 0.773 & 23.06 / 0.660 & 21.27 / 0.515  \\
EDVR~\cite{wang2019edvr}     & 28.72 / 0.805 & 25.98 / 0.706 & 23.36 / 0.600 & 26.34 / 0.771 & 24.45 / 0.667 & 22.31 / 0.534  \\
TecoGAN~\cite{chu2020learning}   & 26.93 / 0.768 & 25.46 / 0.690 & 22.95 / 0.589 & 25.25 / 0.741 & 23.94 / 0.639 & 21.99 / 0.479  \\
FRVSR~\cite{tao2017detail}     & 27.61 / 0.784 & 25.72 / 0.696 & 23.22 / 0.579 & 26.01 / 0.766 & 24.33 / 0.655 & 22.05 / 0.482  \\
RSDN~\cite{isobe2020video}      & 27.66 / 0.768 & 25.48 / 0.679 & 23.03 / 0.579 & 26.58 / 0.781 & 24.06 / 0.650 & 21.29 / 0.483  \\
BasicVSR~\cite{chan2021basicvsr}  & 29.05 / 0.814 & 25.93 / 0.704 & 23.22 / 0.596 & 26.56 / 0.780 & 24.28 / 0.656 & 21.97 / 0.509  \\
IconVSR~\cite{chan2021basicvsr}   & 29.10 / 0.816 & 25.93 / 0.704 & 23.22 / 0.596 & 26.65 / 0.782 & 24.31 / 0.657 & 21.97 / 0.509  \\
MuCAN~\cite{li2020mucan}     & 28.67 / 0.804 & 25.96 / 0.705 & 23.55 / 0.600 & 25.85 / 0.753 & 24.34 / 0.661 & 22.26 / 0.531  \\
COMISR~\cite{li2021comisr}    & 28.40 / 0.809 & 26.47 / 0.728 & 23.56 / 0.599 & 26.43 / 0.791 & 24.97 / 0.701 & 22.35 / 0.509  \\
FTVSR~\cite{qiu2022learning}     & 30.51 / 0.853 & 28.05 / 0.776 & 24.82 / 0.657 & 27.40 / 0.811 & 25.38 / 0.706 & 22.61 / 0.540  \\
FTVSR++~\cite{qiu2023learning}   & \underline{30.70 / 0.857} & \underline{28.28 / 0.783} & \underline{25.07 / 0.668} & \underline{27.72 / 0.820} & \underline{25.61 / 0.713} & \underline{22.85 / 0.652}  \\
\midrule 
{FCVSR (ours)}    & \textbf{30.78 / 0.863}& \textbf{28.36 / 0.788}& \textbf{25.15 / 0.671} & \textbf{27.76 / 0.823}  & \textbf{25.66 / 0.716} & \textbf{22.87 / 0.655} \\
\bottomrule
\end{tabular}
\end{table*}

\subsection{Comparison with State-of-the-Art VSR methods}

The quantitative results of our models for three training-test sets are summarized in Table~\ref{tab_PSNR}. 
It can be observed from Table~\ref{tab_PSNR} that compared with the  State-of-the-Art (SoTA) VSR model, i.e., IA-RT~\cite{xu2024enhancing}, our FCVSR model achieves similar performance at low QP values and higher performance at high QP values in most cases in terms of three quality metrics.

To comprehensively demonstrate the effectiveness of our models, visual comparison results have been provided in Fig.~\ref{vision_RST}, in which example blocks generated by FCVSR models are compared with those produced by IconVSR, BasicVSR++, FTVSR++, MIA-VSR, and IA-RT. It is clear in these examples that our results contain fewer artifacts and finer details compared to other benchmarks.

The results of the model complexity comparison in terms of model parameters, FLOPs, and FPS for all the models tested are provided in Table~\ref{tab_PSNR}. Here, inference speed (FPS) is based on the REDS4 dataset. Among all the VSR methods, our FCVSR-S model is associated with the lowest model complexity based on three complexity measurements. The complexity-performance trade-off can also be illustrated by Fig.~\ref{fig_perfcomplexity}, in which all FCVSR models are all above the Pareto front curve formed by five benchmark methods. This confirms the practicality of the proposed FCVSR models.

Constant rate factor (CRF)-based video compression was also widely used for compressed video super-resolution. To further evaluate our FCVSR model in this compression configuration, following the same procedure as previous works, i.e., FTVSR~\cite{qiu2022learning} and FTVSR++~\cite{qiu2023learning}, we generate the CRF-based compressed videos on the REDS~\cite{nah2019ntire} and Vimeo-90K~\cite {xue2019video} datasets to train our FCVSR model. Eleven state-of-the-art VSR methods, i.e., DUF~\cite{jo2018deep}, EDVR~\cite{wang2019edvr}, TecoGAN~\cite{chu2020learning}, FRVSR~\cite{tao2017detail}, RSDN~\cite{isobe2020video},  BasicVSR~\cite{chan2021basicvsr}, IconVSR~\cite{chan2021basicvsr}, MuCAN~\cite{li2020mucan}, COMISR~\cite{li2021comisr}, FTVSR~\cite{qiu2022learning}, and FTVSR++~\cite{qiu2023learning} were compared and their results were obtained from the FTVSR++~\cite{qiu2023learning} paper. These results and the results of our FCVSR model are reported in Table~\ref{Tab_CRF}. It can be observed from Table~\ref{Tab_CRF} that our FCVSR model achieves the best performance for all CRF factors on the two testing datasets.

Additionally, to further evaluate the super-resolution performance of our FCVSR model on uncompressed videos, we train our FCVSR model on the REDS~\cite{nah2019ntire} and Vimeo-90K~\cite {xue2019video} datasets. Ten state-of-the-art VSR methods, i.e., DUF~\cite{jo2018deep}, EDVR~\cite{wang2019edvr},  COMISR~\cite{li2021comisr}, BasicVSR~\cite{chan2021basicvsr}, IconVSR~\cite{chan2021basicvsr}, TTVSR~\cite{liu2022learning}, FTVSR~\cite{qiu2022learning}, FTVSR++~\cite{qiu2023learning}, MIA-VSR~\cite{zhou2024video} and  IA-RT~\cite{xu2024enhancing} were compared.  The results of DUF~\cite{jo2018deep}, EDVR~\cite{wang2019edvr},  COMISR~\cite{li2021comisr}, BasicVSR~\cite{chan2021basicvsr}, IconVSR~\cite{chan2021basicvsr}, TTVSR~\cite{liu2022learning}, FTVSR~\cite{qiu2022learning}, FTVSR++~\cite{qiu2023learning} were obtained from the FTVSR++~\cite{qiu2023learning} paper, and MIA-VSR~\cite{zhou2024video} and  IA-RT~\cite{xu2024enhancing} were obtained from their own papers. The results of these methods and our FCVSR method were provided in Table~\ref{Tab_uncompressed}. It can be observed from Table~\ref{Tab_uncompressed} that our FCVSR surpasses FTVSR++~\cite{qiu2023learning} in terms of PSNR on the REDS4 dataset and achieves the best PSNR performance on the Vid4 dataset.

\begin{table}[!t]
\centering
\caption{Quantitative Comparison in terms of PSNR (dB) and SSIM on the uncompressed videos of REDS4 and Vid4 datasets. The best and the second-best results are highlighted and underlined.} \label{Tab_uncompressed}
\setlength{\tabcolsep}{5mm}
\fontsize{7}{9}\selectfont
\begin{tabular}{r|c|c}
\toprule 
\multirow{2}{*}{Methods} & REDS4~\cite{nah2019ntire} & Vid4~\cite{xue2019video} \\
& PSNR$\uparrow$ / SSIM$\uparrow$ & PSNR$\uparrow$ / SSIM$\uparrow$ \\
\midrule 
DUF~\cite{jo2018deep}       & 28.63 / 0.825 & 27.38 / 0.832  \\
EDVR~\cite{wang2019edvr}      & 31.09 / 0.880 & 27.85 / 0.850  \\
COMISR~\cite{li2021comisr}    & 29.68 / 0.868 & 27.31 / 0.840  \\
BasicVSR~\cite{chan2021basicvsr}  & 31.42 / 0.890 & 27.96 / 0.855  \\
IconVSR~\cite{chan2021basicvsr}   & 31.67 / 0.895 & 28.04 / 0.857  \\
TTVSR~\cite{liu2022learning}     & 32.12 / 0.901 & 28.40 / \underline{0.864}  \\
FTVSR~\cite{qiu2022learning}     & 31.82 / 0.896 & 28.31 / 0.860  \\
FTVSR++~\cite{qiu2023learning}   & 32.42 / 0.907 & 28.80 / \textbf{0.868}  \\
{MIA-VSR~\cite{zhou2024video}}   & 32.78 / \textbf{0.922}  & 28.20 / 0.851 \\
 {IA-RT~\cite{xu2024enhancing}}    & \textbf{32.90} / \underline{0.914}  & {28.26 / 0.852}  \\
\midrule 
{FCVSR (ours)}   &  32.67 / \underline{0.914}  & \textbf{28.82} / {0.861} \\
\bottomrule
\end{tabular}
\end{table}

\begin{table}[!t]
\caption{Ablation study results for the proposed FCVSR model.} \label{Tab_module}
\setlength{\tabcolsep}{1.40mm}
\fontsize{7}{9}\selectfont
\centering
\begin{tabular}{lcccccccc}
\toprule
{\multirow{2}{*}{{Models}}}  & {\multirow{2}{*} {PSNR(dB)$\uparrow$/SSIM$\uparrow$/VMAF$\uparrow$}} & \multicolumn{1}{c}{Param.$\downarrow$}  & \multicolumn{1}{c} {FLOPs$\downarrow$}  & \multicolumn{1}{c}{FPS$\uparrow$} \\
&   & (M)  & (G)  &  (1/s)\\
\midrule
(v1.1) w/o MGAA  & 25.04 / 0.6615 / 30.62 & 8.25   & 155.30 & 3.43  \\
(v1.2) w/o ME &  25.12 / 0.6641 / 31.63 & 8.49   & 157.91 & 3.62 \\
(v1.3) Flow(Spynet)  &  24.83 / 0.6565 / 28.67 & 6.82  & 129.29  & 4.89  \\
(v1.4) Flow(RAFT)  &  25.07 / 0.6620 / 31.27 & 10.63  & 173.74  & 2.85  \\
(v1.5) DCN   &  25.01 / 0.6598 / 30.79 & 8.79  & 170.84   & 3.02  \\
(v1.6) FGDA  &  25.10 / 0.6631 / 31.74 & 10.45  & 210.64  & 2.10  \\
\midrule
(v2.1) w/o MFFR  &  25.10 / 0.6630 / 31.45 & 8.20  & 159.57 & 3.02  \\
(v2.2) w/o FBE  &  25.16 / 0.6668 / 31.95  & 8.81  & 165.36  & 2.68  \\
(v2.3) w/o FFE & 25.14 / 0.6664 / 31.92 & 8.81  & 165.36 & 2.76  \\
\midrule
(v3.1) w/o $\mathcal{L}_{fc}$  & 25.12 / 0.6652 / 31.85   & 8.81  & 165.36  & 2.39 \\
(v3.2) w/o $\mathcal{L}^{1}_i$ & 25.15 / 0.6676 / 31.92   & 8.81  & 165.36  & 2.39 \\
(v3.3) w/o $\mathcal{L}^{2}_i$ & 25.17 / 0.6682 / 31.97   & 8.81  & 165.36  & 2.39 \\
\midrule
FCVSR  & \textbf{25.20 / 0.6694 / 32.05}   & 8.81  & 165.36  & 2.39 \\
\bottomrule
\end{tabular}
\end{table}

\begin{table}[!t]
\caption{Verify the effectiveness of different decomposition numbers in the MFFR module for our FCVSR model.} \label{Tab_MFFR}
\setlength{\tabcolsep}{4.0mm}
\fontsize{7}{9}\selectfont
\centering
\begin{tabular}{c|cccccccccc}
\midrule[0.2mm]
$Q$ & PSNR(dB)$\uparrow$/SSIM$\uparrow$/VMAF$\uparrow$  & Param.(M)$\downarrow$  & FPS(720P)$\uparrow$ \\
\midrule[0.2mm]
1  &  24.93 / 0.6583 / 29.05  & 8.54    & 2.64     \\ 
2  &  25.02 / 0.6604 / 30.17  & 8.58    & 2.52     \\ 
4  &  25.08 / 0.6628 / 30.56  & 8.66    & 2.42     \\
8  &  25.20 / 0.6694 / 32.05  &  8.81   & 2.39     \\
16 &  25.26 / 0.6702 / 32.22  & 9.11    & 2.14     \\
32 &  \textbf{25.29 / 0.6709 / 32.42} & 9.71   & 1.86  \\
\midrule[0.2mm]
\end{tabular}
\end{table}

\begin{table}[t]
\caption{Ablation study results for different band-pass filters for VSR on REDS4 testing dataset.} \label{Tab_filter}
\setlength{\tabcolsep}{5.0mm}
\fontsize{7}{9}\selectfont
\centering
\begin{tabular}{r|ccccc}
\toprule
Filters  & {PSNR(dB)$\uparrow$ / SSIM$\uparrow$ / VMAF$\uparrow$} \\
\midrule
Ideal         &  25.06 / 0.6668 / 31.78  \\
Butterworth   &  25.13 / 0.6681 / 31.89  \\
Gaussian (ours)   & \textbf{25.20 / 0.6694 / 32.05}  \\
\bottomrule
\end{tabular}
\end{table}

\begin{table}[t]
\caption{Ablation study results for the hyper-parameter $\alpha$ for VSR on REDS4 testing dataset.} \label{Tab_aset}
\setlength{\tabcolsep}{7.40mm}
\fontsize{7}{9}\selectfont
\centering
\begin{tabular}{c|cccc}
\toprule
$\alpha$ & {PSNR(dB)$\uparrow$ / SSIM$\uparrow$ / VMAF$\uparrow$}  \\
\midrule
0.4  &  25.12 / 0.6676 / 31.85    \\ 
0.6  &  25.15 / 0.6686 / 31.98    \\ 
0.8  &  25.18 / 0.6690 / 32.02    \\ 
1.0  &  \textbf{25.20 / 0.6694 / 32.05}    \\
1.2  &  25.16 / 0.6683 / 31.96    \\
1.4  &  25.12 / 0.6671 / 31.78    \\
\bottomrule
\end{tabular}
\end{table}

\subsection{Ablation Study}

To further verify the effectiveness of the main contributions in this work, we have created different model variants in the ablation study, and used the REDS4 dataset (QP = 37) in this experiment.

We first tested the contribution of the MGAA module (and its sub-blocks) by creating the following variants. (v1.1) w/o MGAA - the MGAA module is removed and the features of frames are fused by a concatenation operation and a convolution layer to obtain the aligned features. We have also tested the effectiveness of the Motion Estimator within the MGAA module, obtaining (v1.2) w/o ME - the input neighboring features are directly fed into the MGAC layer without the guidance of motion offsets to generate the aligned features. The MGAA module has also been replaced by other existing alignment modules, including flow-based alignment modules ((v1.3) Spynet~\cite{ranjan2017optical} and (v1.4) RAFT~\cite{teed2020raft}), deformable convolution-based alignment modules ((v1.5) DCN~\cite{zhu2019deformable}), and flow-guided deformable alignment module ((v1.6) FGDA~\cite{chan2022basicvsr++}), to verify the effectiveness of MGAA module. 
The corresponding results are provided in Table~\ref{Tab_module}. It can be seen that our MGAA module surpasses the current SOTA alignment module, i.e., FGDA, 0.10dB PSNR, and reduces 1.64M parameters and 45.28G FLOPs. These results indicate that our MGAA has high performance and efficiency. It benefits from the generation of multiple motion offsets and the design of cascaded adaptive convolutions. Multiple motion offsets with different receptive fields include diverse motion information between frames. Cascaded adaptive convolutions are guided by the multiple motion offsets, driving the model for feature alignment with low complexity, thereby achieving high performance and better flexibility.

The effectiveness of the MFFR module has also been fully evaluated by removing it from the pipe, resulting in (v2.1) w/o MFFR. The contributions of each branch in this module have also been verified by creating (v2.2) w/o FBE - removing the feedback enhancement branch and (v2.3) w/o FFE - disabling the feedforward enhancement branch.  Specifically, for the ``w/o FBE" model, the FBE branch is removed and only remains the FFE branch, the forward enhanced feature $E^{f}_{q}$ as the final enhanced feature $E_q$. For the ``w/o FFE" model, the FFE branch is removed and only remains the FBE branch, the feedback enhanced feature $E^{b}_{q}$ is added with the current band feature $S_{q}$ to obtain the final enhanced feature $E_{q}$. Additionally, to verify the number of decompositions, we set the number of decompositions $Q$ from 1 to 32 in the MFFR module.  The corresponding experimental results are provided in Table~\ref{Tab_MFFR}. It can be found from Table~\ref{Tab_MFFR} that the higher decomposition number, the better performance and higher complexity. We choose the $Q=8$ as a good balance between performance and complexity. Different decomposition numbers can bring significant performance changes and few parameter changes, which proves that the MFFR module has a crucial impact on the SR performance of compressed videos.
In the MFFR module, the band-pass filter in $Decoupler$ directly decides the performance of feature refinement. To evaluate the different band-pass filters for feature refinement, we adopted the Ideal band-pass filter, the Butterworth band-pass filter, and the Gaussian band-pass filter to construct the MFFR module, the corresponding experimental results are provided in Table \ref{Tab_filter}. It can be found in Table~\ref{Tab_filter} that the Gaussian band-pass filter brings the highest performance, which confirms our selection for the Gaussian filter.

The results of these variants and the full FCVSR model have been summarized in Table~\ref{Tab_module}. It can be observed that the full FCVSR model is outperformed by all these model variants in terms of three quality metrics, which confirms the contributions of these key models and their sub-blocks.

\begin{figure}[t]
\centering
\begin{minipage}[b]{0.180\linewidth}
\centering
\centerline{\epsfig{figure=LRQP22_000_00000004.png,width=1.65cm}}
\footnotesize{LR Frame}  
\end{minipage}
\begin{minipage}[b]{0.180\linewidth}
\centering
\centerline{\epsfig{figure=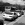,width=1.65cm}}
\footnotesize{LL}  
\end{minipage}
\begin{minipage}[b]{0.180\linewidth}
\centering
\centerline{\epsfig{figure=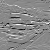,width=1.65cm}}
\footnotesize{LH} 
\end{minipage}
\begin{minipage}[b]{0.180\linewidth}
\centering
\centerline{\epsfig{figure=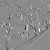,width=1.65cm}}
\footnotesize{HL} 
\end{minipage}
\begin{minipage}[b]{0.180\linewidth}
\centering
\centerline{\epsfig{figure=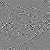,width=1.65cm}}
\footnotesize{HH} 
\end{minipage}

\vspace{3pt}

\centering
\begin{minipage}[b]{0.180\linewidth}
\centering
\centerline{\epsfig{figure=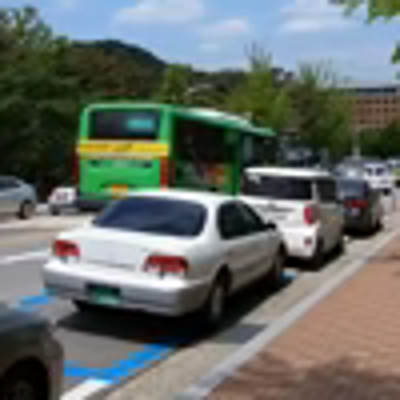,width=1.65cm}}
\footnotesize{Upsampled}  
\end{minipage}
\begin{minipage}[b]{0.180\linewidth}
\centering
\centerline{\epsfig{figure=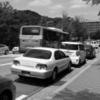,width=1.65cm}}
\footnotesize{LL }  
\end{minipage}
\begin{minipage}[b]{0.180\linewidth}
\centering
\centerline{\epsfig{figure=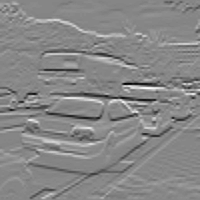,width=1.65cm}}
\footnotesize{LH } 
\end{minipage}
\begin{minipage}[b]{0.180\linewidth}
\centering
\centerline{\epsfig{figure=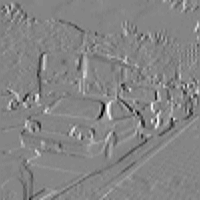,width=1.65cm}}
\footnotesize{HL } 
\end{minipage}
\begin{minipage}[b]{0.180\linewidth}
\centering
\centerline{\epsfig{figure=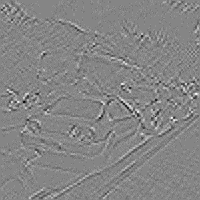,width=1.65cm}}
\footnotesize{HH  } 
\end{minipage}

\vspace{3pt}

\centering
\begin{minipage}[b]{0.180\linewidth}
\centering
\centerline{\epsfig{figure=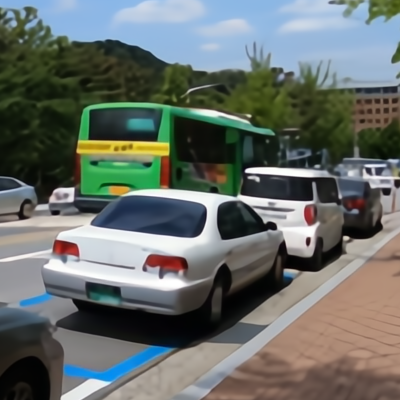,width=1.65cm}}
\footnotesize{SR Frame}  
\end{minipage}
\begin{minipage}[b]{0.180\linewidth}
\centering
\centerline{\epsfig{figure=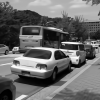,width=1.65cm}}
\footnotesize{LL}  
\end{minipage}
\begin{minipage}[b]{0.180\linewidth}
\centering
\centerline{\epsfig{figure=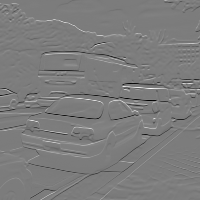,width=1.65cm}}
\footnotesize{LH} 
\end{minipage}
\begin{minipage}[b]{0.180\linewidth}
\centering
\centerline{\epsfig{figure=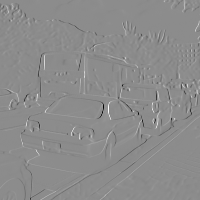,width=1.65cm}}
\footnotesize{HL} 
\end{minipage}
\begin{minipage}[b]{0.180\linewidth}
\centering
\centerline{\epsfig{figure=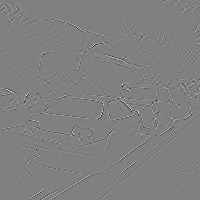,width=1.65cm}}
\footnotesize{HH} 
\end{minipage}

\vspace{3pt}

\begin{minipage}[b]{0.180\linewidth}
\centering
\centerline{\epsfig{figure=GT_000_00000004.png,width=1.65cm}}
\footnotesize{GT}  
\end{minipage}
\begin{minipage}[b]{0.180\linewidth}
\centering
\centerline{\epsfig{figure=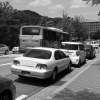,width=1.65cm}}
\footnotesize{LL}  
\end{minipage}
\begin{minipage}[b]{0.180\linewidth}
\centering
\centerline{\epsfig{figure=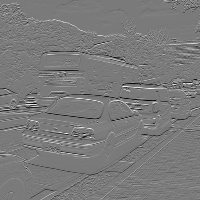,width=1.65cm}}
\footnotesize{LH} 
\end{minipage}
\begin{minipage}[b]{0.180\linewidth}
\centering
\centerline{\epsfig{figure=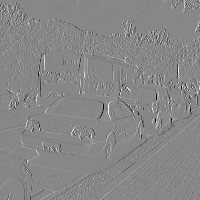,width=1.65cm}}
\footnotesize{HL} 
\end{minipage}
\begin{minipage}[b]{0.180\linewidth}
\centering
\centerline{\epsfig{figure=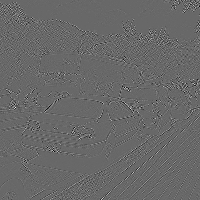,width=1.65cm}}
\footnotesize{HH} 
\end{minipage}
\caption{Visualization of subbands of compressed LR frame, Upsampled SR frame, SR frame (ours) and Ground-Truth (GT).}\label{visionDWT}
\end{figure}

Finally, to test the contribution of the proposed frequency-aware contrastive loss, we retrained our FCVSR model separately by removing the $\mathcal{L}_{fc}$ (v3.1) or its high/low frequency terms, $\mathcal{L}_i^1$ (v3.2) and $\mathcal{L}_i^2$ (v3.3), respectively, resulting in three additional variants as shown in Table~\ref{Tab_module}. It can be observed that the proposed $\mathcal{L}_{fc}$ (and its high/low frequency sub losses) does consistently contribute to the final performance. Additionally, we verify the hyper-parameter $\alpha$ in our loss and the corresponding experimental results were provided in Table~\ref{Tab_aset}. It can be found that when $\alpha$ was set as 1.0, the super-resolution performance reached its optimal. It indicates that spatial-based loss and frequency-based loss have the same weight, which is more conducive to model optimization. To further analyze the design of our loss function, we provide visualizations of each frequency band extracted by the 2D discrete wavelet transform, the visual results of the compressed LR frame, the upsampled SR frame, our SR frame, and the Ground-Truth (GT) are illustrated in Fig.~\ref{visionDWT}. 
It is noted that for the low-frequency subbands, the SR result is similar to the GT frame and the upsampled SR frame. For high-frequency subbands, SR results are similar to GT frame and dissimilar to the upsampled SR frames, which verifies the motivation for our proposed frequency-aware contrastive loss.

\section{Conclusion}  \label{CON}

In this paper, we proposed a frequency-aware video super-resolution network, FCVSR, for compressed video content, which consists of a new motion-guided adaptive alignment (MGAA) module for improved feature alignment and a novel multi-frequency feature refinement (MFFR) module that enhances the fine detail recovery. A frequency-aware contrastive loss is also designed for training the proposed FCVSR network with optimal super-resolution performance. We have conducted a comprehensive comparison experiment and adequate ablation study to evaluate the performance of the proposed method and its primary contributions. Due to its superior performance and relatively low computational complexity, we believe this work makes a strong contribution to the research field of video super-resolution and is suitable for various application scenarios.

\small
\bibliographystyle{IEEEtran}
\bibliography{IEEEabrv,freqCVSR}

\vfill

\end{document}